%% file: egpaper_for_review.tex
\documentclass[10pt,twocolumn,letterpaper]{article}

\usepackage{cvpr}
\usepackage{times}
\usepackage{epsfig}
\usepackage{graphicx}
\usepackage{amsmath}
\usepackage{amssymb}
\usepackage{caption}
\usepackage{subcaption}
\usepackage{booktabs}
\usepackage{enumitem}
\usepackage{float}
\usepackage[misc]{ifsym}

\newcommand{\eat}[1]{}

\usepackage[pagebackref=true,breaklinks=true,letterpaper=true,colorlinks,bookmarks=false]{hyperref}

\cvprfinalcopy % *** Uncomment this line for the final submission

\def\cvprPaperID{1938} % *** Enter the CVPR Paper ID here

\ifcvprfinal\fi
\begin{document}

%%%%%%%%% TITLE
\title{CLEVR-Ref+: Diagnosing Visual Reasoning with Referring Expressions}

\author{Runtao Liu$^1$, Chenxi Liu$^2$$^{(\textrm{\Letter})}$, Yutong Bai$^3$, Alan Yuille$^2$\\
$^1$Peking University \quad $^2$Johns Hopkins University \quad $^3$Northwestern Polytechnical University\\
{\tt\small runtao219@gmail.com \quad cxliu@jhu.edu \quad ytongbai@gmail.com \quad alan.l.yuille@gmail.com}
% For a paper whose authors are all at the same institution,
% omit the following lines up until the closing ``}''.
% Additional authors and addresses can be added with ``\and'',
% just like the second author.
% To save space, use either the email address or home page, not both
}

\maketitle
%\thispagestyle{empty}

%%%%%%%%% ABSTRACT
\begin{abstract}
Referring object detection and referring image segmentation are important tasks that require joint understanding of visual information and natural language. Yet there has been evidence that current benchmark datasets suffer from bias, and current state-of-the-art models cannot be easily evaluated on their intermediate reasoning process. To address these issues and complement similar efforts in visual question answering, we build CLEVR-Ref+, a synthetic diagnostic dataset for referring expression comprehension. The precise locations and attributes of the objects are readily available, and the referring expressions are automatically associated with functional programs. The synthetic nature allows control over dataset bias (through sampling strategy), and the modular programs enable intermediate reasoning ground truth without human annotators.

In addition to evaluating several state-of-the-art models on CLEVR-Ref+, we also propose IEP-Ref, a module network approach that significantly outperforms other models on our dataset. In particular, we present two interesting and important findings using IEP-Ref: (1) the module trained to transform feature maps into segmentation masks can be attached to any intermediate module to reveal the entire reasoning process step-by-step; (2) even if all training data has at least one object referred, IEP-Ref can correctly predict no-foreground when presented with false-premise referring expressions. To the best of our knowledge, this is the first direct and quantitative proof that neural modules behave in the way they are intended.\footnote{All data and code concerning CLEVR-Ref+ and IEP-Ref have been released at \url{https://cs.jhu.edu/~cxliu/2019/clevr-ref+}}
\end{abstract}

%%%%%%%%% BODY TEXT
\input{intro}
\input{related}
\input{dataset}
\input{exp}
\input{conc}

\vspace{-0.5cm}
\paragraph{Acknowledgments}
This research is support by NSF award CCF-1317376 and ONR N00014-12-1-0883.

{\small
\bibliographystyle{ieee}
\bibliography{egbib}
}

\input{supp}

\end{document}

%% file: intro.tex
\vspace{-0.4cm}
\section{Introduction}

There has been significant research interest in the joint understanding of vision and natural language. 
While image captioning \cite{DBLP:conf/cvpr/KarpathyL15,DBLP:conf/cvpr/DonahueHGRVDS15,DBLP:journals/corr/MaoXYWY14a,DBLP:conf/aaai/LiuMSY17} focuses on generating a sentence with image being the only input, visual question answering (VQA) \cite{DBLP:conf/iccv/AntolALMBZP15, DBLP:conf/nips/GaoMZHWX15, DBLP:conf/cvpr/ZhuGBF16} and referring expressions (REF) \cite{DBLP:conf/cvpr/MaoHTCY016,DBLP:conf/cvpr/HuXRFSD16} require comprehending both an image and a sentence, before generating an output.
In this paper, we focus on referring expressions, which is to identify the particular objects (in the form of segmentation mask or bounding box) in a given scene from natural language.

In order to study referring expressions, various datasets have been proposed \cite{DBLP:conf/cvpr/MaoHTCY016,DBLP:conf/eccv/YuPYBB16,DBLP:conf/emnlp/KazemzadehOMB14}.
These are real-world images annotated by crowdsource workers.
The advantage of these datasets is that they, to a certain extent, reflect the complexity and nuances of the real world.
Yet inevitably, they also have limitations.
First, they usually exhibit strong biases that may be exploited by the models \cite{DBLP:conf/naacl/CirikMB18}.
Roughly speaking, this means simply selecting the salient foreground object (i.e., discarding the referring expression) will yield a much higher baseline than random.
This casts doubts on the true level of understanding within current REF models.
Second, evaluation can only be conducted on the final segmentation mask or bounding box, but not the intermediate step-by-step reasoning process.
For example, for the referring expression ``Woman to the left of the red suitcase'', a reasonable reasoning process should be first find all suitcases in the image, then identify the red one among them, finally segment the woman to its left.
Clearly this requires significantly more high-quality annotations, which are currently unavailable and hard to collect.

\begin{figure*}[t]
\captionsetup[subfigure]{labelformat=empty, font=footnotesize}
\centering
\includegraphics[width=0.33\linewidth]{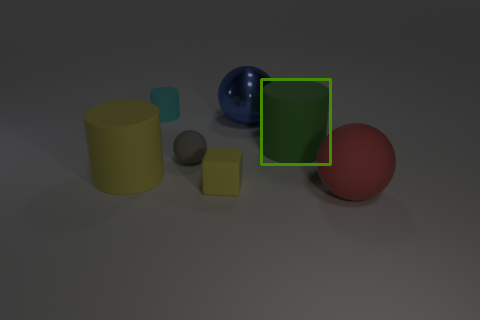}
\includegraphics[width=0.33\linewidth]{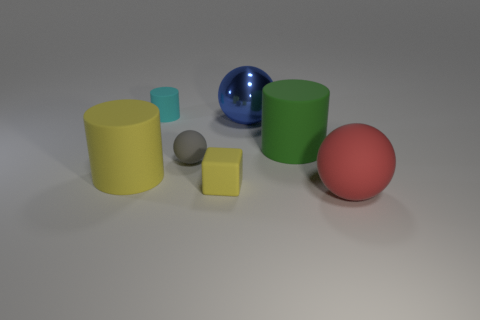}
\includegraphics[width=0.33\linewidth]{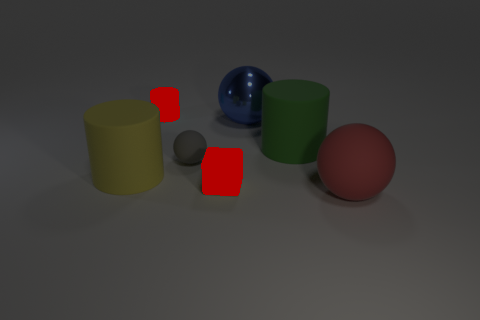}
\begin{subfigure}{0.45\linewidth}
\smallskip
\caption{\textit{The big thing(s) that are behind the second one of the big thing(s) from front and to the right of the first one of the large sphere(s) from left}}
\end{subfigure}
\begin{subfigure}{0.08\linewidth}
\caption{}
\end{subfigure}
\begin{subfigure}{0.45\linewidth}
\smallskip
\caption{\textit{Any other things that are the same size as the fifth one of the thing(s) from right}}
\end{subfigure}
\vspace{-0.35cm}
\caption{Examples from our CLEVR-Ref+ dataset. We use the same scenes as those provided in CLEVR \cite{DBLP:conf/cvpr/JohnsonHMFZG17}. Instead of asking questions about the scene, we ask the model to either return one bounding box (as illustrated on the left) or return a segmentation mask (could potentially be multiple objects; illustrated on the right) based on the given referring expression. }
\label{fig:fig1}
\vspace{-0.1cm}
\end{figure*}

To address these concerns and echo similar efforts in visual question answering (i.e., CLEVR \cite{DBLP:conf/cvpr/JohnsonHMFZG17}), we propose CLEVR-Ref+, a synthetic diagnostic dataset  for referring expressions.
The advantage of using a synthetic dataset is that we have full control over the scene, and dataset bias can be minimized by employing a uniform sampling strategy.
Also, the referring expressions are now automatically annotated with the true underlying reasoning process, so a step-by-step analysis becomes much more plausible.

We make much effort in constructing CLEVR-Ref+ to make sure it is well adapted and applicable to the referring expression task.
First, we turn the original questions in CLEVR into their corresponding referring expression format. 
Second, we change the output space from textual answers (in the form of a word) to referred objects (in the form of segmentation mask or bounding box).
Third, we analyzed statistics from real-world REF datasets and found that there are some common types of referring expressions (e.g., ``The second sphere from left'') that are not included in CLEVR templates.
In our CLEVR-Ref+, we add support for these types of expressions to better match the variety of referring expressions used in real world.

We tested several state-of-the-art referring expression models on our CLEVR-Ref+ dataset.
This includes both those designed for referring segmentation \cite{DBLP:conf/iccv/LiuLSYLY17} and detection \cite{DBLP:conf/cvpr/YuTBB17,DBLP:conf/cvpr/YuLSYLBB18}.
In addition to evaluating the overall IoU and accuracy as previous datasets, we can now do a more detailed breakdown and analysis in terms of sub-categories.
For example, we found that it is especially hard for the models to understand ordinality.
This could point to important research directions in the future.

Besides diagnosing these existing models, we also propose IEP-Ref, a Neural Module Network \cite{DBLP:conf/cvpr/AndreasRDK16} solution based on IEP \cite{DBLP:conf/iccv/JohnsonHMHFZG17}. 
% We adapt it from producing textual answers to generating the masks of referred objects from the last module output. 
Experiments show that the IEP-Ref model achieved excellent performance on CLEVR-Ref+ with its explicit, step-by-step functional program and module network execution engine, suggesting the importance of compositionality.
Very interestingly, we found that the module trained on translating the last module output to segmentation mask is general, and can produce excellent human-interpretable segmentation masks when attached to intermediate module outputs, revealing the entire reasoning process.
We believe ours is the first to show clean visualization of the visual reasoning process carried out by neural module networks, as opposed to gradient norms \cite{DBLP:conf/iccv/JohnsonHMHFZG17} or soft attention maps \cite{mascharka2018transparency, DBLP:conf/eccv/HuADS18}. 

In sum, our paper makes the following contributions:
\begin{itemize}
\item We construct CLEVR-Ref+, a synthetic diagnostic dataset for referring expression tasks that complements existing real-world datasets.
\item We test and diagnose several state-of-the-art referring expression models on CLEVR-Ref+, including our proposed IEP-Ref that explicitly captures compositionality.
\item The segmentation module trained in IEP-Ref can be trivially plugged in all intermediate steps in the module network to produce excellent segmentation masks that clearly reveal the network's reasoning process.
\end{itemize}

%% file: related.tex
\section{Related Works}

\subsection{Referring Expressions}

Referring expressions are sentences that refer to specific objects in an image.
Understanding referring expressions has important applications in robotics and human-computer interaction.
In recent years, many deep learning models have been developed for this task.

Several works focused on detection, i.e. returning one bounding box containing the referred object.
\cite{DBLP:conf/cvpr/MaoHTCY016,DBLP:conf/cvpr/HuXRFSD16} adapted image captioning for this task by scoring each bounding box proposal with a generative captioning model. 
\cite{DBLP:conf/eccv/RohrbachRHDS16} learned the alignment between the description and image region by reconstructing the description using an attention mechanism.
\cite{DBLP:conf/eccv/YuPYBB16, DBLP:conf/eccv/NagarajaMD16} studied the importance of context for referring expressions.
\cite{DBLP:conf/cvpr/LuoS17} used a discriminative comprehension model to improve referring expression generation.
\cite{DBLP:conf/cvpr/YuTBB17} showed additional gain by incorporating reinforcement learning.
\cite{DBLP:conf/cvpr/HuRADS17, DBLP:conf/cvpr/YuLSYLBB18} used learned parser and module networks to better match the structured semantics.

There are also works focusing on segmentation, i.e. returning the segmentation mask.
\cite{DBLP:conf/eccv/HuRD16} used FCN feature concatenated with LSTM feature to produce pixel-wise binary segmentation.
\cite{DBLP:conf/iccv/LiuLSYLY17} used a convolutional LSTM in addition to the language-only LSTM to facilitate propagation of intermediate segmentation beliefs.
\cite{li2018referring, DBLP:conf/eccv/Margffoy-TuayPB18} improved upon \cite{DBLP:conf/iccv/LiuLSYLY17} by making more architectural improvements.

\eat{
In our paper, we chose \cite{DBLP:conf/cvpr/YuTBB17,DBLP:conf/cvpr/YuLSYLBB18} among detection methods and \cite{DBLP:conf/iccv/LiuLSYLY17} among segmentation methods to be tested on our CLEVR-Ref+ dataset.
}

\begin{table*}
\centering
\caption{Examples of converting questions to referring expressions. }
\label{tab:question_to_refexp}
\vspace{-0.2cm}
\begin{tabular}{lp{7cm}p{7cm}}
\toprule
{\bf Category} & {\bf Question (CLEVR)}  & {\bf Referring Expression (CLEVR-Ref+)}  \\
\midrule
 Basic & How many cyan cubes are there?  & The cyan cubes. \\
 
 Spatial Relation & Are there any green cylinders to the left of the brown sphere? & The green cylinders to the left of the brown sphere. \\
 
 AND Logic & How many green spheres are both in front of the red cylinder and left to the yellow cube? & The green spheres that are both in front of the red cylinder and left to the yellow cube. \\
 
 OR Logic & Are there any cylinders that are either purple metal objects or small red matte things?  & Cylinders that are either purple metal objects or small red matte things. \\
 
 Same Relation & Are there any other things that have the same size as the red sphere?  & The things/objects that have the same size as the red sphere. \\
 
 Compare Integer & Are there more brown shiny objects behind the large rubber cylinder than gray blocks? & -\\
 
 Comparison      & Does the small ball have the same color as the small cylinder in front of the big sphere? & - \\
 
\bottomrule
\end{tabular}
\vspace{-0.2cm}
\end{table*}

\subsection{Dataset Bias and Diagnostic Datasets}

In visual question answering, despite exciting models being proposed and accuracy on benchmark datasets being steadily improved, there has been serious concern over the dataset bias problem \cite{DBLP:conf/cvpr/ZhangGSBP16, DBLP:conf/cvpr/GoyalKSBP17}, meaning that models may be heavily exploiting the imbalanced distribution in the training/testing data.
More recently, \cite{DBLP:conf/naacl/CirikMB18} showed that dataset bias also exists in referring expression datasets \cite{DBLP:conf/cvpr/MaoHTCY016,DBLP:conf/emnlp/KazemzadehOMB14,DBLP:conf/eccv/YuPYBB16}.
For example, \cite{DBLP:conf/naacl/CirikMB18} reported that the performance when discarding the referring expression and basing solely on the image is significantly higher than random.
Ideally the dataset should be unbiased so that the performance faithfully reflect the model's true level of understanding.
But this is very hard to control when working with real-world images and human-annotated referring expressions.

A possible solution is to use synthetic datasets. 
Indeed this is the path taken by CLEVR \cite{DBLP:conf/cvpr/JohnsonHMFZG17}, a diagnostic dataset for VQA.
There, objects are placed on a 2D plane and only have a small number of choices in terms of shape, color, size, and material. 
The question-answer pairs are also synthesized using carefully designed templates.
Together with a uniform sampling strategy, this design can mitigate dataset bias and reveal the model's ability to understand compositionality.
We construct our CLEVR-Ref+ dataset by re-purposing CLEVR towards the referring expression task. 

Several approaches now achieve near-perfect accuracy on CLEVR \cite{DBLP:conf/iccv/JohnsonHMHFZG17,DBLP:conf/iccv/HuARDS17,DBLP:conf/aaai/PerezSVDC18,DBLP:conf/nips/SantoroRBMPBL17,mascharka2018transparency,hudson2018compositional,DBLP:conf/eccv/HuADS18}.
In addition to reporting the VQA accuracy, they typically try to interpret the visual reasoning process through visualization.
However, the quality of these visualizations does not match the high VQA accuracy.
We suspect the primary reason is that the domain these models are trained for (i.e. a textual answer) is different from the domain these models are diagnosed on (i.e. attention over the image).
Fortunately, in referring expressions these two domains are very much interchangeable\eat{, and we will show in Section~\ref{sec:exp} how we could explain the visual reasoning process through excellent, human-interpretable binary segmentation results}.

Note that CLEVR was also adapted towards referring expression in \cite{DBLP:conf/eccv/HuADS18}, but they focused on facilitating VQA, instead of introducing extensions (Section~\ref{sec:module-additions}), evaluating state-of-the-art models (Section~\ref{sec:models}), and directly facilitating the diagnosis of visual reasoning (Section~\ref{sec:inspection}).

%% file: dataset.tex
\section{The CLEVR-Ref+ Dataset}

CLEVR-Ref+ uses the exact same scenes as CLEVR (70K images in train set, 15K images in validation and test set), and every image is associated with 10 referring expressions.
Since CLEVR is a VQA dataset, we began by changing the questions to referring expressions (Section~\ref{sec:question-to-refexp}), and the answers to referred objects (Section~\ref{sec:answer-to-obj}).
We then made important additions to the set of modules (Section~\ref{sec:module-additions}) as well as necessary changes to the sampling procedure (Section~\ref{sec:generation}).
Finally, we made the distinction whether more than one object is being referred (Section~\ref{sec:multi-single}).

\subsection{From Question to Referring Expression}
\label{sec:question-to-refexp}

Templates are provided in CLEVR so that questions and the functional programs associated with them can be generated at the same time.
We notice that in many cases, part of the question is indeed a referring expression, as we need to first identify objects of interest before asking about their property (e.g. color or number).
In Table~\ref{tab:question_to_refexp} we provide examples of how we change question templates into their corresponding referring expression templates, usually by selecting a subset.
The associated functional programs are also adjusted accordingly.
For example, for ``How many'' questions, we simply remove the \texttt{Count} module at the end.

The original categories ``Compare Integer'' and ``Comparison'' were about comparing properties of two groups of referred objects, so they do not contribute additional referring expression patterns. 
Therefore they are not included in the templates for CLEVR-Ref+.

\subsection{From Answer to Referred Objects}
\label{sec:answer-to-obj}

In referring expressions, the output is no longer a textual answer, but a bounding box or segmentation mask. 

Since we know the exact 3D locations and properties of objects in the scene, we can follow the ground truth functional program associated with the referring expression to identify which objects are being referred.
In fact we can do this not only at the end (also available in real-world datasets), but also at every intermediate step (not available in real-world datasets).
This will become useful later when we do step-by-step inspection and evaluation of the visual reasoning process.

After finding the referred objects, we project them back to the image plane to get the ground truth bounding box and segmentation mask.
This automatic annotation was done through rendering with the software Blender.
For occluded objects, only the visible part is treated as ground truth.

\begin{table}[t]
\centering
\caption{Frequent category and words in RefCOCO+ \cite{DBLP:conf/eccv/YuPYBB16}.}
\label{tab:stat-refcoco+}
\begin{tabular}{lll}
\toprule
{\bf Category} & {\bf Example words} & {\bf Frequency}\\
\midrule
object   & shirt,head,chair,hat,pizza  & 63.66\%\\
human    & man,woman,guy,girl,person & 42.54\%\\
color    & white,black,blue,red,green&  38.76\%\\
spatial  & back,next,behind,near,up&  23.86\%\\
animal   & zebra,elephant,horse,bear&  15.36\%\\
attribute& big,striped,small,plaid,long&  10.55\%\\
action   & standing,holding,looking&  10.34\%\\
{\bf ordinal}  & {\bf closest,furthest,first,third} &  {\bf 5.797\%} \\
compare  & smaller,tallest,shorter,older&  5.247\%\\
{\bf visible}  & {\bf fully visible,barely seen} &  {\bf 4.639\%} \\
\bottomrule
\end{tabular}
\end{table}

\subsection{Module Additions}
\label{sec:module-additions}

We hope the referring expressions that we generate are representative of those used in the real world.
However, since the task is no longer the same, we suspect that there may be some frequent referring patterns missing in the templates directly inherited from CLEVR.
To this end, we analyzed statistics from a real-world referring expression dataset, RefCOCO+ \cite{DBLP:conf/eccv/YuPYBB16}, as shown in Table~\ref{tab:stat-refcoco+}.

We began by sorting the words in these referring expressions by their frequency. 
Then, starting with the most frequent word, we empirically cluster these words into categories.
Not surprisingly, nouns that represent object or human are the most common.
However, going down the list, we found that the ``ordinal'' (e.g. ``The second woman from left'') and ``visible'' (e.g. ``The barely seen backpack'') categories recall more than 10\% of all sentences, but are not included in the existing templates.
Moreover, it is indeed possible to define them using a computer program, because there is no ambiguity in meaning.
We add these two new modules into the CLEVR-Ref+ function catalog\eat{, as shown in Figure~\ref{fig:ordinal-visible}}.

\eat{
\begin{figure}[t]
\centering
\includegraphics[width=\linewidth]{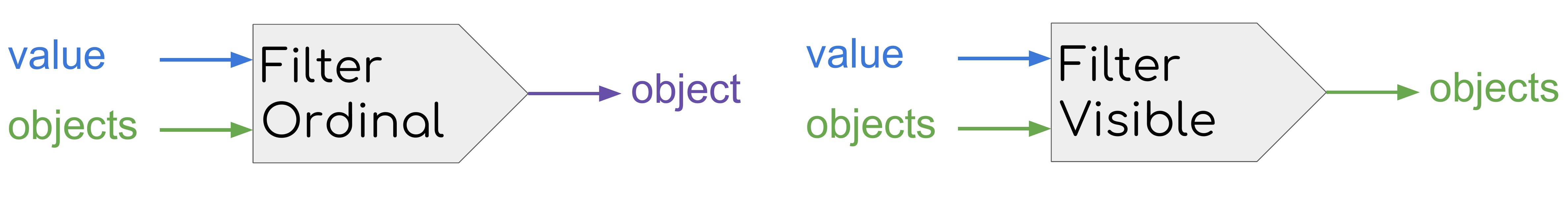}
\caption{We introduce two new modules, \texttt{Ordinal} and \texttt{Visible}, to better cover the variety of referring patterns.}
\label{fig:ordinal-visible}
\end{figure}
}

In a functional program, these two modules may be inserted whenever color, material, size, or shape is being described. 
As an example, ``the red sphere'' may be equivalently described as ``the third sphere from left'' or ``the partially visible red object''.
In our dataset, we define an object to be \emph{partially visible} if foreground objects' mask occupies more than 20\% of its bounding box area. 
For an object to be \emph{fully visible}, this value must be exactly 0. 
We do not describe visibility when there is an ambiguous case (i.e. this value is between 0 and 0.2) in the scene.

\begin{table*}
\centering
\caption{Referring object detection and referring image segmentation results on CLEVR-Ref+. We evaluated three existing models, as well as IEP-Ref which we adapted from its VQA counterpart.}
\label{tab:accuracy-iou}
\vspace{-0.2cm}
\begin{tabular}{l | c | ccc | cc | c | cc}
\toprule
& Basic & \multicolumn{3}{c|}{Spatial Relation} & \multicolumn{2}{c|}{Logic} & \\
             & 0-Relate   & 1-Relate & 2-Relate & 3-Relate & AND & OR & Same & Accuracy & IoU \\
\midrule
SLR \cite{DBLP:conf/cvpr/YuTBB17}         & 0.627   & 0.569 & 0.570 & 0.584 & 0.594 & 0.701 & 0.444 & 0.577 & - \\
MAttNet \cite{DBLP:conf/cvpr/YuLSYLBB18}      & 0.566   & 0.623 & 0.634 & 0.624 & 0.723 & 0.737 & 0.454 & 0.609 & - \\
\midrule
\midrule
RMI \cite{DBLP:conf/iccv/LiuLSYLY17}         & 0.822   & 0.713 & 0.736 & 0.715 & 0.585 & 0.679 & 0.251 & - & 0.561\\
IEP-Ref (GT)    & 0.928   & 0.895 & 0.908 & 0.908 & 0.879 & 0.881 & 0.647 & - & 0.816\\
IEP-Ref (700K prog.) & 0.920 & 0.884 & 0.902 & 0.898 & 0.860 & 0.869 & 0.636 & - & 0.806\\
IEP-Ref (18K prog.) & 0.907 & 0.858 & 0.874 & 0.862 & 0.829 & 0.847 & 0.605 & - & 0.782\\
IEP-Ref (9K prog.) & 0.910 & 0.858 & 0.847 & 0.811 & 0.778 & 0.791 & 0.626 & - & 0.760\\
\bottomrule
\end{tabular}
\end{table*}

\subsection{Generation Procedure}
\label{sec:generation}

Generating a referring expression for a scene is conceptually simple and intuitive. 
The process may be summarized as the following few steps:
\begin{enumerate}
\setlength\itemsep{0em}
\item Randomly choose a referring expression family\footnote{A referring expression family contains a template for constructing functional programs and several text templates that provide multiple ways of expressing these programs in natural language. }.
\item Randomly choose a text template from this family.
\item Follow the functional program and select random values when encountering template parameters\footnote{For instance, left/right/front/behind; big/small; metal/rubber.}.
\item Reject when certain criteria fail, that is, the sampled referring expression is inappropriate for the given scene; return when the entire functional program follows through.
\end{enumerate}

We largely follow the generation procedure of CLEVR, with a few important changes:
\begin{itemize}
\setlength\itemsep{0em}
\item To balance the number of referring expressions across different categories (those listed in  Table~\ref{tab:question_to_refexp}), we double the probability of being sampled in categories with a small number of referring expression families. 
\item When describing the attributes for a set of objects, we do not use \texttt{Ordinal} and \texttt{Visible} at the same time. 
This is because referring an object as ``The second partially visible object from left'' seems too peculiar and rare, and there usually exists more natural alternatives.
\item Originally when describing the attributes for a set of objects, four fair coins were flipped to determine whether color, material, size, shape will be included.
As a result, usually multiple attributes are selected, and a very small number of objects survive these filters.
We empirically found that this makes it quite easy for the system to select the correct object simply from the attributes that directly describe the target object(s).

To remedy this, we first enumerate all possible combinations of these attributes, and calculate how many objects will survive for each possibility. 
We then uniformly sample from these possible number of survivors, before doing another uniform sampling to find the combination of attributes. 
This will ensure a larger variance in terms of number of objects after each set of filtering, and prevent near-degenerate solutions.
\item At the end of the functional program, we verify if at least one object is being referred; reject otherwise.
\end{itemize}

\subsection{Multi-Object and Single-Object Referring}
\label{sec:multi-single}

As explained in Section~\ref{sec:generation}, each referring expression in CLEVR-Ref+ may refer to one or more objects in the scene.
We believe this is the more general setting, and models should have the flexibility to handle various number of objects being referred.
This is already handled and supported by referring image segmentation systems.
However, we notice that detection based systems are usually designed to return a single object instead of multiple objects, presumably because this was how the detection datasets \cite{DBLP:conf/cvpr/MaoHTCY016,DBLP:conf/eccv/YuPYBB16} were created.
As a result, for detection based methods, we evaluate on the subset of CLEVR-Ref+ where only a single object is referred.
This subset contains a total of 222,569 referring expressions (32\% of the entire dataset).

%% file: exp.tex
\section{Experiments}
\label{sec:exp}

\begin{figure*}[t]
\centering
\includegraphics[width=0.93\linewidth]{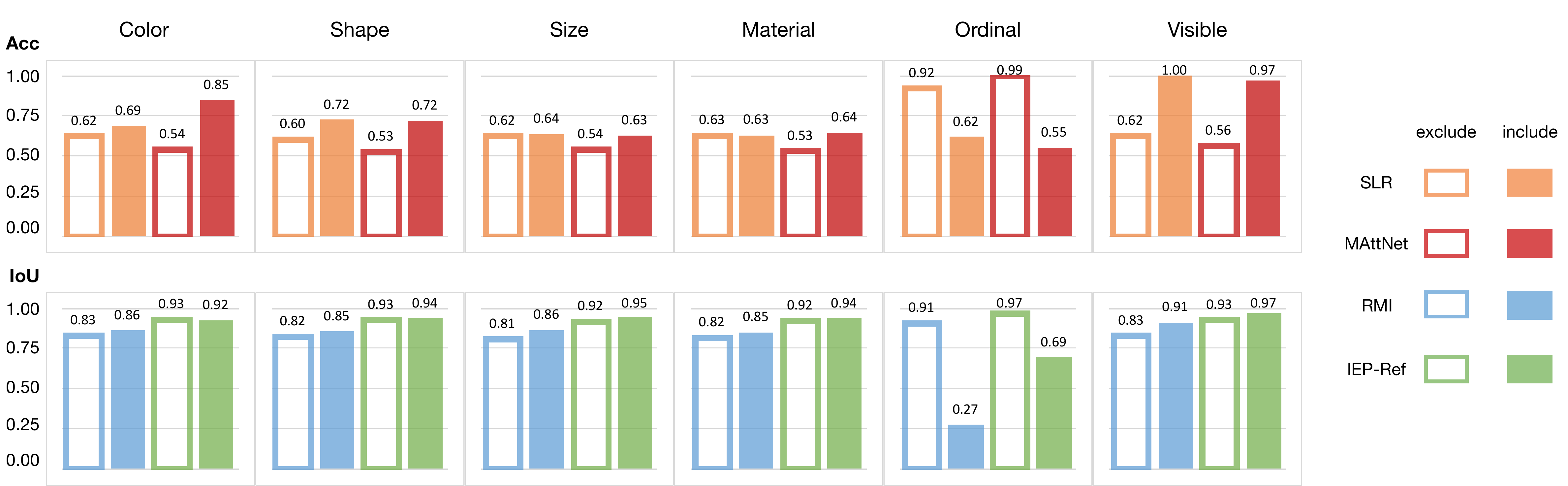}
\vspace{-0.1cm}
\caption{Analyzing the basic referring ability of different models. ``Include'' means the average performance if a module is involved in the referring process. ``Exclude'' means otherwise. As a result, high ``exclude'' and low ``include'' performance suggests that this module is more challenging to learn, and vice versa. }
\label{fig:basic_ability}
\vspace{-0.5cm}
\end{figure*}

\subsection{Models and Implementation Details}
\label{sec:models}

In all models we resize the input image to $320 \times 320$ to set up a fair comparison.
Publicly available code for these models are used with minimum change to adapt to our CLEVR-Ref+ dataset.
The following referring expression models are studied and tested:

\vspace{-0.4cm}
\paragraph{Speaker-Listener-Reinforcer (SLR) \cite{DBLP:conf/cvpr/YuTBB17}}
This is a \textbf{detection} model that includes a generative model (speaker), a discriminative model (listener), as well as a reinforcement learning component that makes further improvement.
%#####vl_sim model#####
%lr 4e-4
%epochs 26 epochs(180000 iterations, 222569 examples, 32 batch_size)
%batch_size 32
%#####main model#####
%lr 4e-4
%epochs 7 (50165 iterations, 222569 examples, 32 batch_size)
%batch_size 32
Before training the main model, the visual-language similarity model needs to be trained first.
We use Adam optimizer \cite{DBLP:journals/corr/KingmaB14}, learning rate 4e-4, batch size 32 for both the visual-language similarity model and the main model.

\vspace{-0.4cm}
\paragraph{MAttNet \cite{DBLP:conf/cvpr/YuLSYLBB18}}
This is also a \textbf{detection} model, that uses three modular networks to capture the subject, location, and relationship features respectively. 
A soft attention mechanism is used to return the overall score of a candidate region.
%lr 4e-4
%epochs 6 (80,000 iterations, 222569 examples, 15 batch_size)
%batch_size 15
We use learning rate 4e-4 and batch size 15.

\vspace{-0.4cm}
\paragraph{Recurrent Multimodal Interaction (RMI) \cite{DBLP:conf/iccv/LiuLSYLY17}}
This is a \textbf{segmentation} model.
In addition to concatenating the referring expression LSTM embedding with the image features, RMI also used a convolutional LSTM to facilitate propagation of segmentation beliefs when reading in the referring expression word-by-word.
%lr 0.00025 = 2.5*1e-4 (with degration)
%epochs 2.5 (330,000 iterations, 70w examples, 3 batch_size)
%batch_size 3
%input_size 320x320
%loss_compute size 320x320
We use Adam optimizer, learning rate 2.5e-4, batch size 3, and weight decay 5e-4.

\vspace{-0.4cm}
\paragraph{IEP-Ref}
This is a \textbf{segmentation} model that we adapt from IEP \cite{DBLP:conf/iccv/JohnsonHMHFZG17}, which was designed for VQA.
The idea is to use a LSTM program generator to translate the referring expression into a structured series of modules, each of which is parameterized by a small CNN.
By executing this dynamically constructed neural network (with a special \texttt{Segment} module at the end; see supplementary material for its architecture), IEP-Ref imitates the underlying visual reasoning process.
For input visual features, we use the last layer of the \texttt{conv4} stage of ResNet101 \cite{DBLP:conf/cvpr/HeZRS16} pretrained on ImageNet \cite{DBLP:conf/cvpr/DengDSLL009}, which is of size $1024 \times 20 \times 20$. 
Following \cite{DBLP:conf/iccv/JohnsonHMHFZG17}, this part is not finetuned. 
We tried three settings that use 9K/18K/700K ground truth programs to train the LSTM program generator (Adam optimizer, learning rate 5e-4, batch size 64; 20,000 iterations for the 9K setting, 32,000 iterations for the 18K and 700K setting).
The accuracies of the predicted programs are 0.873, 0.971, 0.993 respectively.
For the fourth setting, we simply use the ground truth program\footnote{This is our default IEP-Ref setting unless otherwise specified.}. 
The execution engine is trained for 30 epochs using learning rate 1e-4 and Adam optimizer.

\subsection{Results and Analysis}
\label{sec:analysis}

\begin{figure}[t]
\centering
\includegraphics[width=0.95\linewidth]{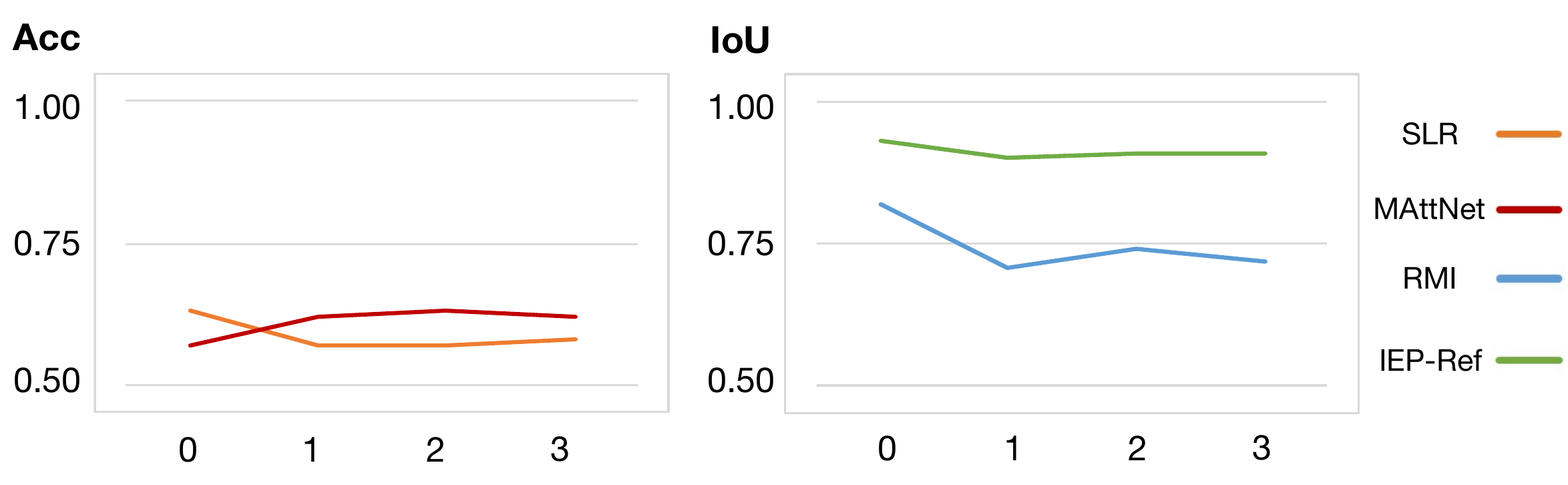}
\vspace{-0.2cm}
\caption{Analyzing the spatial reasoning ability of different models. Horizontal axis is the number of spatial relations.}
\label{fig:0123hop}
\vspace{-0.2cm}
\end{figure}

\subsubsection{Overall Evaluation}
\vspace{-0.1cm}

The experimental results are summarized in Table~\ref{tab:accuracy-iou}.
Detection models are evaluated by accuracy (i.e. whether the prediction selects the correct bounding box among given candidates), where MAttNet performs favorably against SLR. 
Segmentation models are evaluated by Intersection over Union (IoU), where IEP-Ref performs significantly better than RMI.
This suggests the importance to model compositionality within the referring expression. 
We now present a more detailed analysis of various aspects. 

\vspace{-0.25cm}
\subsubsection{Basic Referring Ability}
\vspace{-0.1cm}

Here we start with the easiest form: referring by direct description of object attributes (e.g., ``The big blue sphere'').
Concretely, this corresponds to the ``0-Relate'' subset.

In CLEVR-Ref+, there are totally 6 types of attributes that may help us locate specific objects: color, size, shape, material, ordinality, and visibility. 
In Figure~\ref{fig:basic_ability} we show the average detection accuracy/segmentation IoU of various methods on ``0-Relate'' referring expressions that either contain or not contain a specific type of module.

Among detection models, we found that accuracy is higher when the referring expression contains descriptions of color, shape, and visibility.
A reasonable conjecture is that these concepts are easier to learn compared with the others.
However, for segmentation, the performance gaps between ``exclude'' and ``include'' are not as significant.

Though it is unclear which concept is the easiest to learn, there seems little dispute that ordinality is the hardest. 
In particular, for RMI, IoU is 0.91 if the expression does not require ordinality and 0.27 when it does. 
Other models do not suffer as much, but also experience significant drops.
We suspect this is because ordinality requires the global context, whereas the others are local properties.

\begin{figure}[t]
\centering
\includegraphics[width=0.95\linewidth]{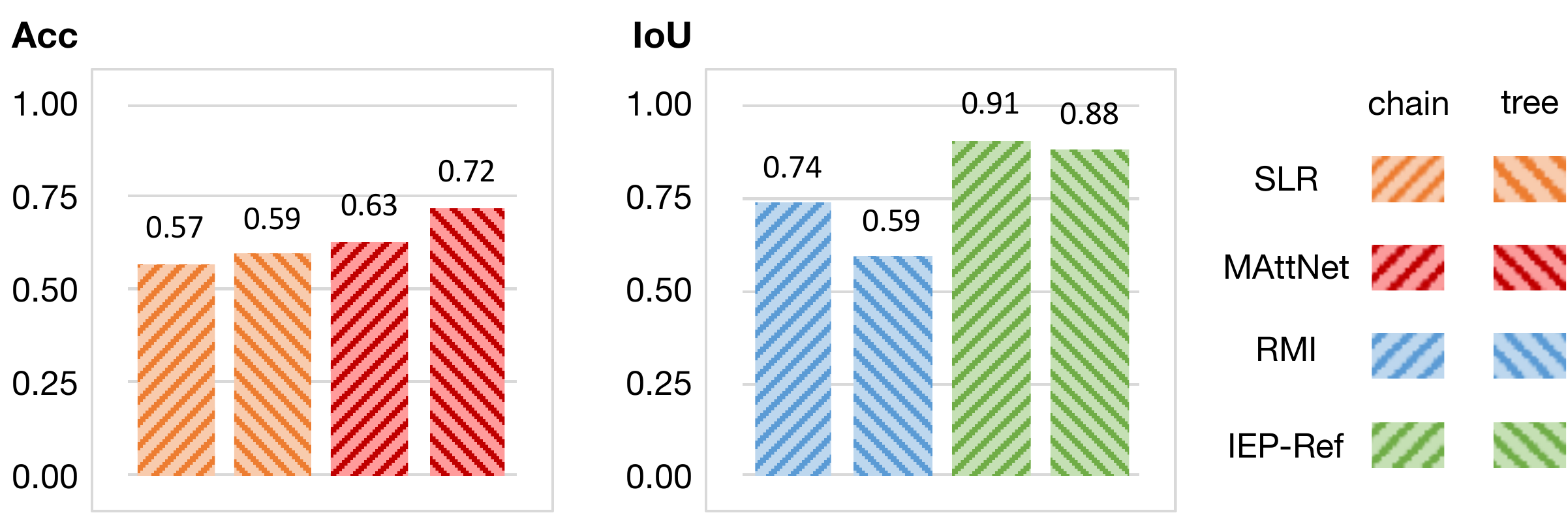}
\vspace{-0.2cm}
\caption{Effect of reasoning topology (Chain vs. Tree) on referring detection or segmentation performance.}
\label{fig:2hop-singleand}
\vspace{-0.2cm}
\end{figure}

\begin{figure}[t]
\centering
\includegraphics[width=0.95\linewidth]{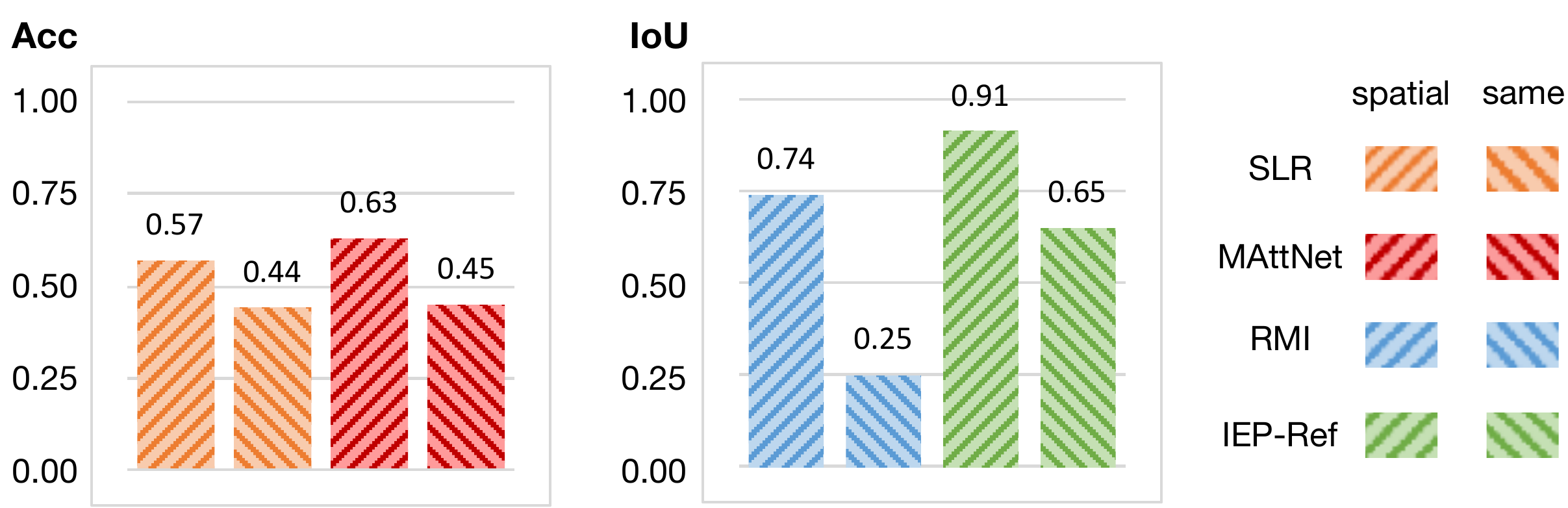}
\vspace{-0.2cm}
\caption{Effect of relation type (Spatial vs. Same) on referring detection or segmentation performance.}
\label{fig:2hop-samerelate}
\vspace{-0.2cm}
\end{figure}

\begin{figure*}[t]
\centering
\includegraphics[width=0.95\linewidth]{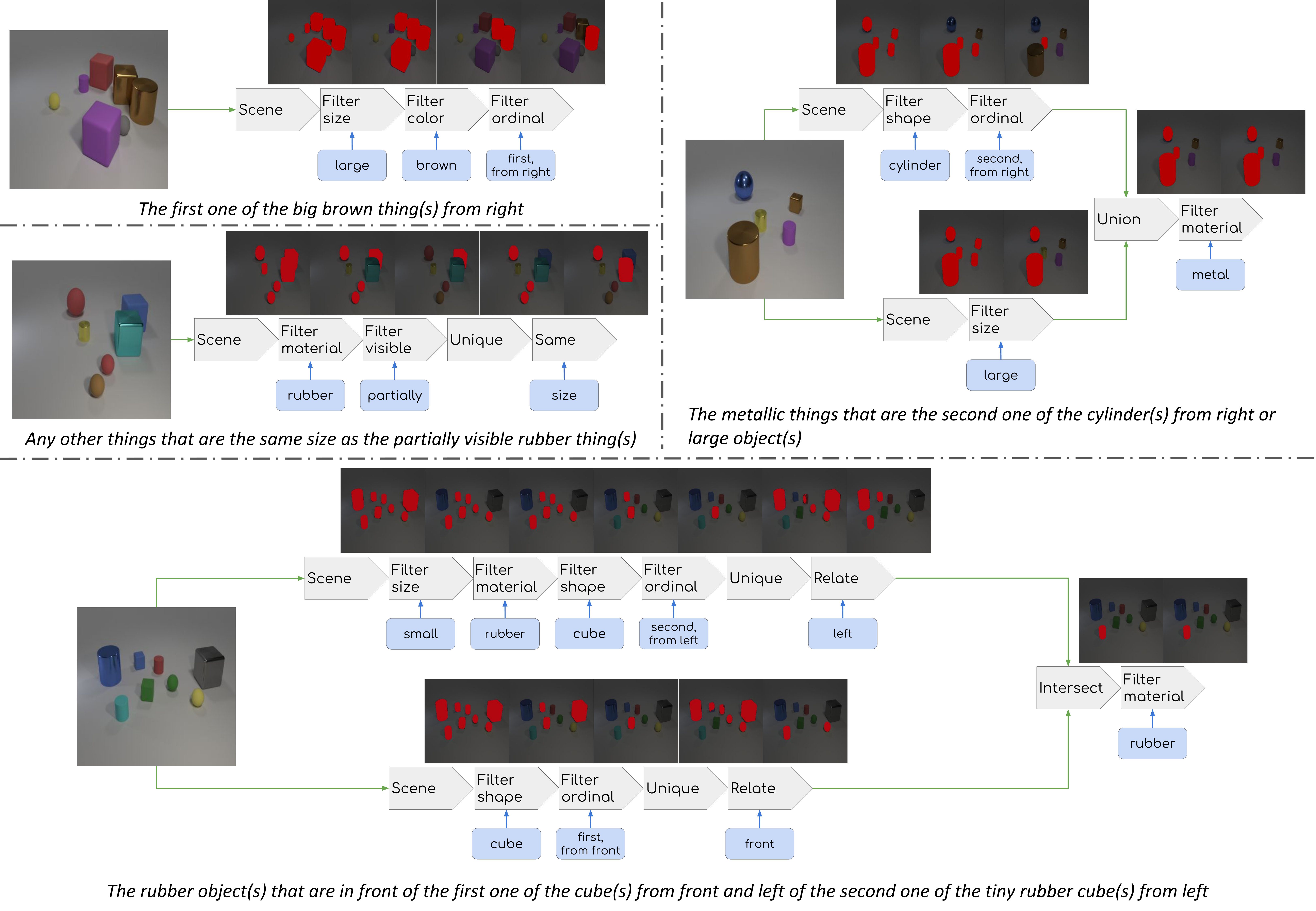}
\caption{Four examples (two chain structures, two tree structures) of step-by-step inspection of IEP-Ref visual reasoning.}
\label{fig:step-by-step}
\end{figure*}

\begin{figure*}[t]
\centering
\includegraphics[width=0.95\linewidth]{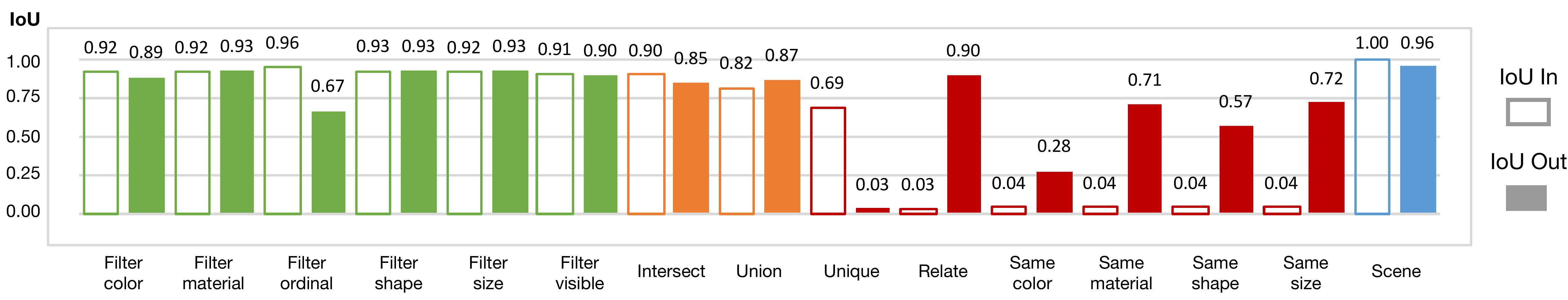}
\caption{Average IoU going into/out of each IEP-Ref module on CLEVR-Ref+ validation set. Note that here IoU is not only computed at the end, but also all intermediate steps. This shows that IoU remains high throughout visual reasoning. The large differences in modules marked in dark red are discussed in text.}
\label{fig:module-before-after}
\vspace{-0.2cm}
\end{figure*}

\vspace{-0.25cm}
\subsubsection{Spatial Reasoning Ability}
\label{sec:spatial}
\vspace{-0.1cm}

Other than directly describing the attributes, it is also common to refer to an object by its spatial location.
Here we diagnose whether referring expression models can understand (potentially multiple steps of) relative spatial relationship, for example ``The object that is left to the red cube''.
In Table~\ref{tab:accuracy-iou}, this corresponds to the ``\{0, 1, 2, 3\}-Relate'' columns.
Results are shown in Figure~\ref{fig:0123hop}.

In general, we observe a small drop when referring expressions start to include spatial reasoning. 
However, there does not seem to be significant difference among referring expressions that require 1, 2, 3 steps of spatial reasoning. 
This seems to suggest that once the model has grasped spatial reasoning, there is little trouble in successfully applying it multiple times.

\vspace{-0.25cm}
\subsubsection{Different Reasoning Topologies}
\vspace{-0.1cm}

There are two referring expression topologies in CLEVR-Ref+: chain-structured and tree-structured.
Intuitively, a chain structure has a single reasoning path to follow, whereas a tree structure requires following two such paths before merging. 
In Figure~\ref{fig:2hop-singleand} we compare performance on referring expressions with two sequential spatial relationships vs. one on each branch joined with AND. 
These two templates have roughly the same length and complexity, so the comparison focuses on topology. 

Though not consistent among the four models, tree-structured referring expressions are generally harder than chain-structured ones.
This agrees with the findings in \cite{DBLP:conf/cvpr/JohnsonHMFZG17}. 

\vspace{-0.25cm}
\subsubsection{Different Relation Types}
\vspace{-0.1cm}

There are two kinds of relationships in CLEVR-Ref+. 
One is spatial relationship that includes phrases like ``left of'', ``right of'', ``in front of'', ``behind'' (discussed in Section~\ref{sec:spatial}).
The other is same-attribute relationship that requires recognizing and memorizing particular attributes of another object, e.g. ``The large block(s) that have the same color as the metal sphere''.

In Figure~\ref{fig:2hop-samerelate} we study whether the relation type will make a difference in performance.
We compare the ``2-Relate'' column with the ``Same'' column in Table~\ref{tab:accuracy-iou}, again because they have roughly the same length and complexity.
All models perform much worse on the same-attribute relationship type, suggesting that this is a hard concept to grasp.
Similar to ordinality, same-attribute requires global context.

\subsection{Step-By-Step Inspection of Visual Reasoning}
\label{sec:inspection}

All the results discussed in Section~\ref{sec:analysis} are about the endpoint of the visual reasoning process.
We argue that in order to trust the predictions made by the referring expression system, it is also important to make sure that the intermediate reasoning steps make sense.
CLEVR-Ref+ is suitable because: (1) the semantics of the referring expressions is modularized, and (2) the referring ground truth at all intermediate steps can be obtained automatically (i.e. no human annotators needed).

In training our IEP-Ref model, there is always a \texttt{Segment} module at the end, transforming the 128-channel feature map into a 1-channel segmentation mask. 
When testing, we simply attach the trained \texttt{Segment} module to the output of all intermediate modules. 
This is possible because all modules have the same number of input channels and output channels (128). 
This technique would not help in the VQA setting, because there the ending modules (e.g. \texttt{Count}, \texttt{Equal}) discard the spatial dimensions needed for visualization.

We found that this technique works quite well. 
In Figure~\ref{fig:step-by-step} we provide four qualitative examples with various topologies and modules. 
We notice that all modules are performing their intended functionality, except the \texttt{Unique} module\footnote{It is supposed to simply carry over the previously referred object, yet from what we observe, its behavior is most similar to selecting the complement of the previously referred object, though this is far from consistent.}. 
Yet after one more module, the segmentation mask becomes normal again.
The quantitative analysis in Figure~\ref{fig:module-before-after} confirms this observation: on average, IoU drops by 0.66 after each \texttt{Unique} module; but IoU significantly increases after each \texttt{Same} or \texttt{Relate} module, and these are the only modules that may come after \texttt{Unique} according to the templates. 
We conjecture that the network has learned some mechanism to treat \texttt{Unique} as the ``preprocessing'' step of the \texttt{Same} and \texttt{Relate} functionalities.

\subsection{False-Premise Referring Expressions}

\begin{figure}[t]
\centering
\includegraphics[width=0.95\linewidth]{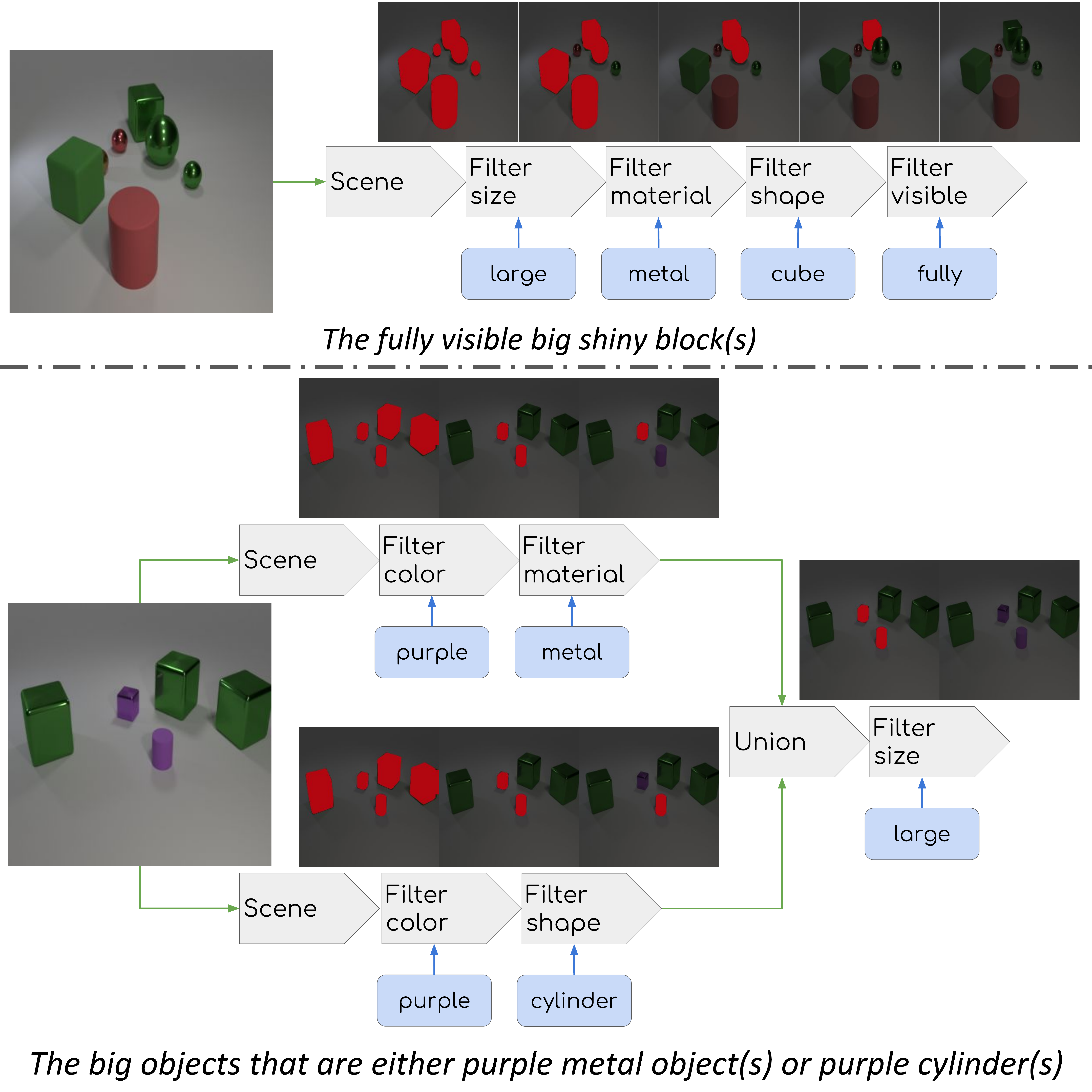}
\vspace{-0.2cm}
\caption{Our IEP-Ref model can correctly handle false-premise referring expressions even if they do not appear during training.}
\label{fig:fake-sent}
\vspace{-0.3cm}
\end{figure}

In reality, referring expression systems may face all kinds of textual input, and not all of them will make sense. 
When presented with a referring expression that makes false assumptions (e.g. ``The red sphere'' when there is no sphere in the scene), the system should follow through as much as it can, and be robust enough to return zero foreground at the end. 
We test IEP-Ref's ability to deal with these false-premise referring expressions (c.f. \cite{DBLP:conf/emnlp/RayCBBP16}).
Note that no such expressions appear during training. 

We generate 10,000 referring expressions that refer to zero object at the end. 
Qualitatively (see Figure~\ref{fig:fake-sent}), it is reassuring to see that intermediate modules are correctly doing their jobs, and a no-foreground prediction is made at the final step.
Quantitatively, IEP-Ref predicts $0$ foreground pixel more than $1/4$ of the time, and $\leq 8$ foreground pixels more than $1/3$ of the time.

\eat{
\subsection{Novel Compositions}

To further test the models' generalization ability, we also conducted experiments on the Compositional Generalization Test (CoGenT) data provided by CLEVR.
Here models are trained on objects with only a subset of all combinations, and then tested on another subset of combinations. 
We report these results in the supplementary material. 
}

%% file: conc.tex
\section{Conclusion}

In this paper, we build the CLEVR-Ref+ dataset to complement existing ones for referring expressions. 
By choosing a synthetic setup, the advantage is that dataset bias can be minimized, and the ground truth visual reasoning process is readily available.
We evaluated several state-of-the-art referring object detection and referring image segmentation models on CLEVR-Ref+.
In addition, we propose the IEP-Ref model, which uses a module network approach and outperforms competing methods by a large margin. 
Detailed analysis are conducted to identify the strengths and weaknesses of these models.
In particular, we found that ordinality and the same-attribute relationship seem to be the most difficult concepts to grasp.

Besides the correctness of the final segmentation mask, the correctness of the reasoning process is also important. 
We discovered that IEP-Ref provides an easy and natural way of revealing this process: simply attach the \texttt{Segment} module to each intermediate step. 
Our quantitative evaluation shows a high IoU at intermediate steps as well, proving that the neural modules have indeed learned the job they are supposed to do.
Another evidence is that IEP-Ref can correctly handle false-premise referring expressions.

Going forward, we are interested to see whether these findings will transfer and inspire better models on real data.

%% file: supp.tex
\clearpage
\begin{center}
    {\Large \bf Supplementary Material \par}
    \vspace*{24pt}
\end{center}
\appendix

In this supplementary material, we begin by providing network architecture details of IEP-Ref to supplement Section~4.1 of the main paper.
We then provide more analysis of the four models' performance on CLEVR-Ref+, to supplement Section~4.2 of the main paper.
Finally, we show more qualitative examples (referring expression and ground truth box/mask) from CLEVR-Ref+. 

\section{Network Architectures in IEP-Ref}

\begin{table}[b]
    \centering
    \begin{tabular}{|c|c|}
    \hline
    Layer & Output size \\
    \hline 
    Input image & $3 \times 320 \times 320$ \\
    ResNet101 \cite{DBLP:conf/cvpr/HeZRS16} conv4\_6 & $1024 \times 20 \times 20$ \\
    Conv($3 \times 3, 1024 \rightarrow 128$) & $128 \times 20 \times 20$ \\
    ReLU & $128 \times 20 \times 20$ \\
    Conv($3 \times 3, 128 \rightarrow 128$) & $128 \times 20 \times 20$ \\
    ReLU & $128 \times 20 \times 20$ \\
    \hline
    \end{tabular}
    \caption{Network architecture for the \textbf{Preprocess} module. }
    \label{tab:preprocess}
\end{table}

\begin{table}[b]
    \centering
    \begin{tabular}{|c|c|c|}
         \hline 
         Index & Layer & Output size \\
         \hline
         (1) & Previous module output & $128 \times 20 \times 20$ \\
         (2) & Conv($3\times 3, 128 \rightarrow 128$) & $128 \times 20 \times 20$ \\
         (3) & ReLU & $128 \times 20 \times 20$ \\
         (4) & Conv($3\times 3, 128 \rightarrow 128$) & $128 \times 20 \times 20$ \\
         (5) & Residual: Add (1) and (4) & $128 \times 20 \times 20$ \\
         (6) & ReLU & $128 \times 20 \times 20$ \\
         \hline
    \end{tabular}
    \caption{Network architecture for the \textbf{Unary} modules.}
    \label{tab:unary}
\end{table}

\begin{table}[t]
    \centering
    \begin{tabular}{|c|c|c|}
         \hline 
         Index & Layer & Output size \\
         \hline
         (1) & Previous module output & $128 \times 20 \times 20$ \\
         (2) & Previous module output & $128 \times 20 \times 20$ \\
         (3) & Concatenate (1) and (2) & $256 \times 20 \times 20$ \\
         (4) & Conv($1\times 1, 256 \rightarrow 128$) & $128 \times 20 \times 20$ \\
         (5) & ReLU & $128 \times 20 \times 20$ \\
         (6) & Conv($3\times 3, 128 \rightarrow 128$) & $128 \times 20 \times 20$ \\
         (7) & ReLU & $128 \times 20 \times 20$ \\
         (8) & Conv($3\times 3, 128 \rightarrow 128$) & $128 \times 20 \times 20$ \\
         (9) & Residual: Add (5) and (8) & $128 \times 20 \times 20$ \\
         (10) & ReLU & $128 \times 20 \times 20$ \\
         \hline
    \end{tabular}
    \caption{Network architecture for the \textbf{Binary} modules.}
    \label{tab:binary}
\end{table}

\begin{table}[t]
    \centering
    \begin{tabular}{|c|c|}
    \hline
    Layer & Output size \\
    \hline 
    Previous module output & $128 \times 20 \times 20$ \\
    Unary module & $128 \times 20 \times 20$ \\
    Conv($1 \times 1, 128 \rightarrow 128$) & $128 \times 20 \times 20$ \\
    ReLU & $128 \times 20 \times 20$ \\
    Bilinear upsample & $128 \times 320 \times 320$ \\
    Conv($1 \times 1, 128 \rightarrow 128$) & $128 \times 320 \times 320$ \\
    ReLU & $128 \times 320 \times 320$ \\
    Conv($1 \times 1, 128 \rightarrow 32$) & $32 \times 320 \times 320$ \\
    ReLU & $32 \times 320 \times 320$ \\
    Conv($1 \times 1, 32 \rightarrow 4$) & $4 \times 320 \times 320$ \\
    ReLU & $4 \times 320 \times 320$ \\
    Conv($1 \times 1, 4 \rightarrow 1$) & $1 \times 320 \times 320$ \\
    \hline
    \end{tabular}
    \caption{Network architecture for the \texttt{Segment} module. }
    \label{tab:postprocess}
\end{table}

In Figure~7 of the main paper, we listed all modules used in our IEP-Ref model (except \texttt{Segment}). 
In IEP-Ref, each of these modules is parameterized with a small fully convolutional network and belongs to one of the following 4 categories:
\begin{itemize}
    \item \textbf{Preprocess}: This component maps the image to the feature tensor. Its output is the input to the \texttt{Scene} module. See Table~\ref{tab:preprocess} for the network architecture.
    \item \textbf{Unary}: This includes the \texttt{Scene}, \texttt{Filter\_X}, \texttt{Unique}, \texttt{Relate}, \texttt{Same\_X} modules. It transforms one feature tensor to another. See Table~\ref{tab:unary} for the network architecture.
    \item \textbf{Binary}: This includes the \texttt{And} and \texttt{Or} modules. It transforms two feature tensors to one. See Table~\ref{tab:binary} for the network architecture. 
    \item \textbf{Postprocess}: This only includes the \texttt{Segment} module. It transforms the 128-channel feature tensor to a 1-channel segmentation mask. See Table~\ref{tab:postprocess} for the network architecture.
\end{itemize}

Network architectures for \textbf{Preprocess}, \textbf{Unary}, \textbf{Binary} are directly inherited from IEP \cite{DBLP:conf/iccv/JohnsonHMHFZG17}.

\section{More Model Analysis on CLEVR-Ref+}
\subsection{Number of Objects in a Scene}
\begin{figure}[t]
\centering
\includegraphics[width=0.95\linewidth]{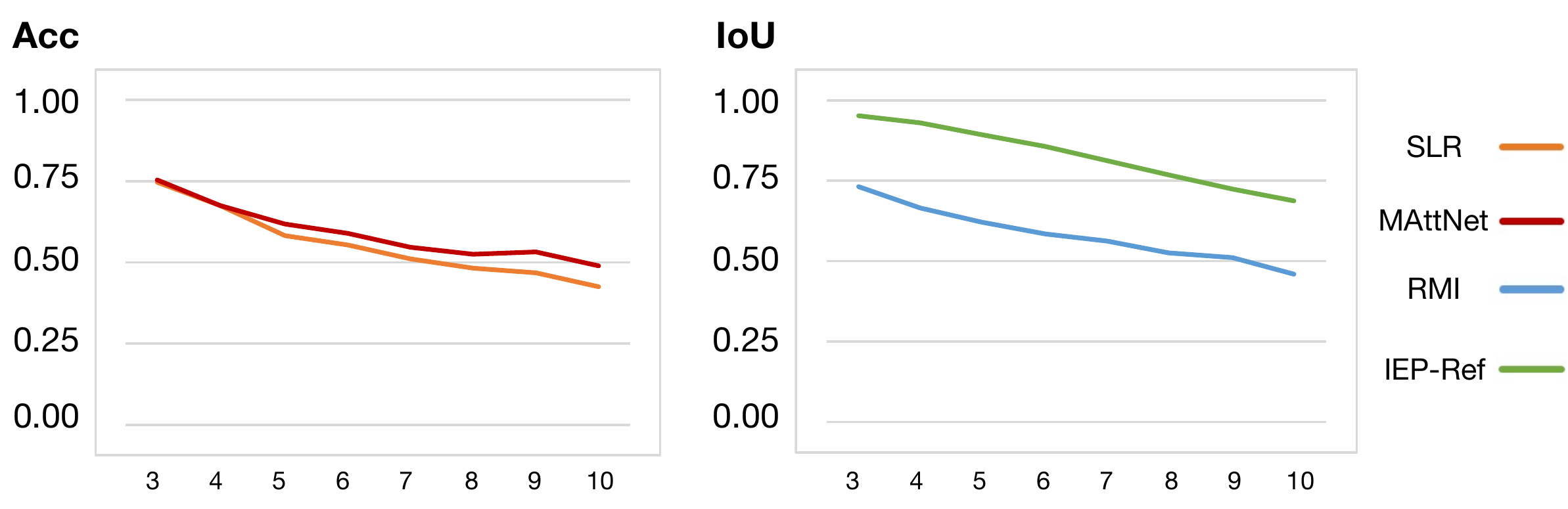}
\caption{Effect of number of objects in a scene on referring detection or segmentation performance.}
\label{fig:num-objects}
\end{figure}

We suspect that the more objects in a scene, the harder for the model to carry out the referring reasoning steps.
In Figure~\ref{fig:num-objects} we plot the performance of each model with respect to the number of objects in a scene.
All models drop in performance when the number of objects increases, suggesting that the models tend to struggle when dealing with too many objects. 

\subsection{Schedule of Acquiring Reasoning Abilities}

\begin{figure}[t]
\centering
\includegraphics[width=0.95\linewidth]{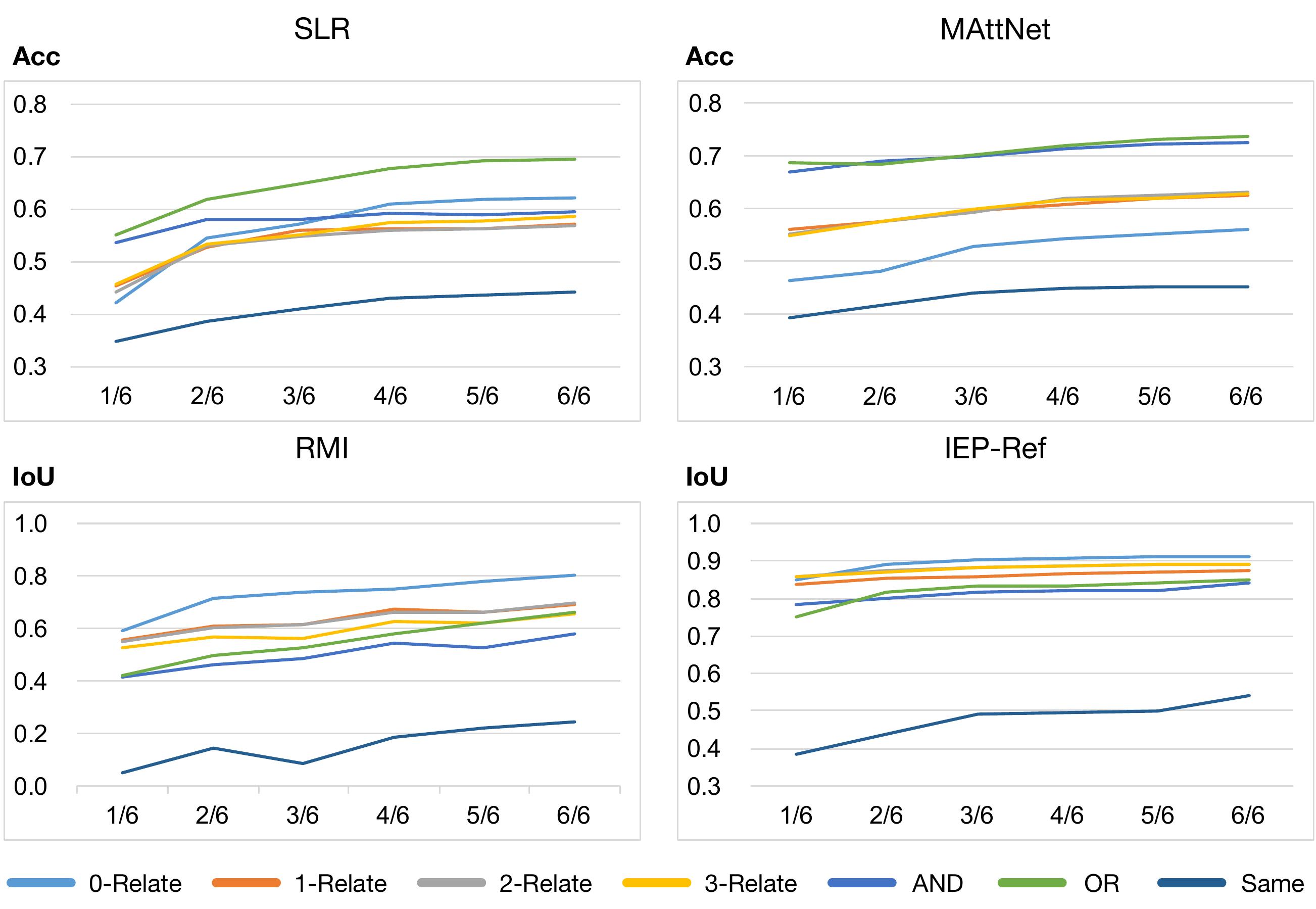}
\caption{Performance across different referring expression categories throughout training. We inspect the performance every 1/6 of the entire training iterations. }
\label{fig:diff-checkpt}
\end{figure}

We are interested to see if throughout the training process, the network exhibit a schedule of acquiring various reasoning abilities (e.g. spatial reasoning, logic etc). 
From Figure~\ref{fig:diff-checkpt}, it seems that no such schedule was developed, and performance steadily increase across different referring expression categories. 
This may be due to the random sampling during training, instead of active learning (c.f. \cite{DBLP:conf/cvpr/MisraGFHGM18}).

\subsection{Novel Compositions}

\begin{figure}[b]
\centering
\includegraphics[width=0.95\linewidth]{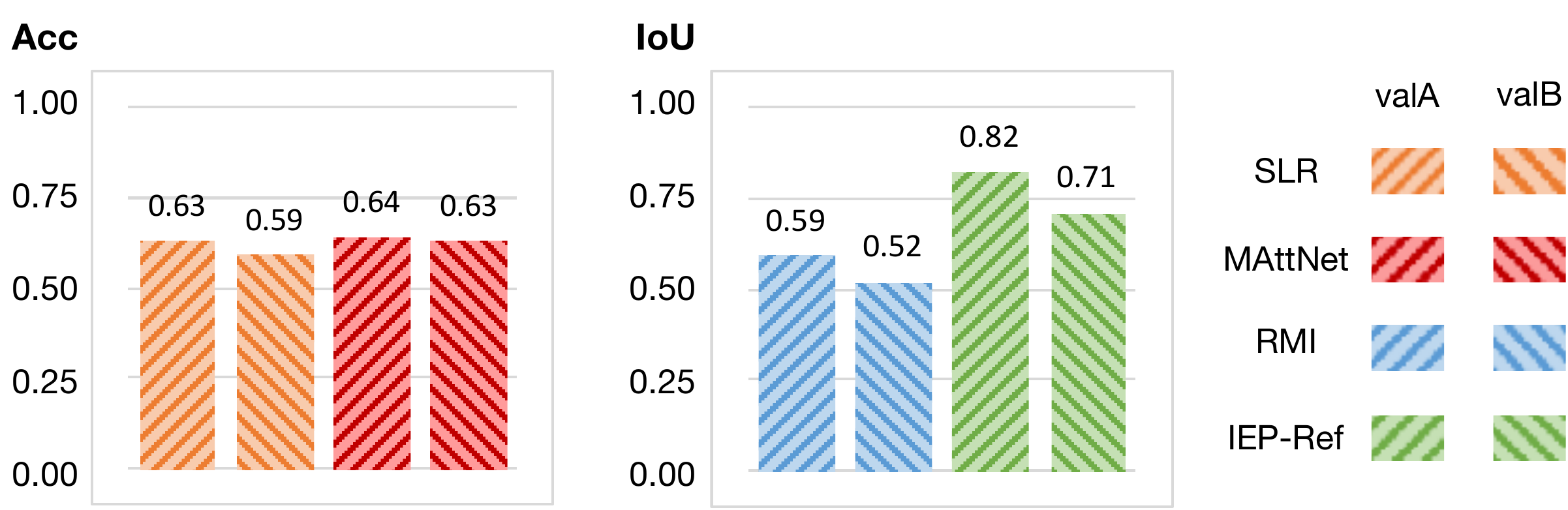}
\caption{Different models' performance on \textit{valA} and \textit{valB} of the CLEVR CoGenT data.}
\label{fig:novel-comp}
\end{figure}

To further test the models' generalization ability, we also conducted experiments on the Compositional Generalization Test (CoGenT) data provided by CLEVR \cite{DBLP:conf/cvpr/JohnsonHMFZG17}.
Here models are trained on objects with only a subset of all combinations, and then tested on both the same subset of combinations (\textit{valA}) and another subset of combinations (\textit{valB}).
Results are summarized in Figure~\ref{fig:novel-comp}.
We see a very small gap for detection models, suggesting that they have learned compositionality to generalize well. 
The gap for segmentation models, on the other hand, is larger. 

\section{More Data Examples from CLEVR-Ref+}

The remaining pages show random images, referring expressions, and the referring ground truth from our CLEVR-Ref+ dataset.
In particular, we choose at least one example from each referring expression category (the 7 middle columns in Table 3 of the main paper).
We show both detection ground truth (Figure~\ref{fig:more-det}) and segmentation ground truth (Figure~\ref{fig:more-seg}).

\input{img_visual.tex}

%% file: img_visual.tex
\begin{figure*}[hbtp]
\centering
\begin{tabular}[c]{cc}

\begin{subfigure}[t]{0.49\linewidth}
  \centering
  \includegraphics[width=0.49\linewidth]{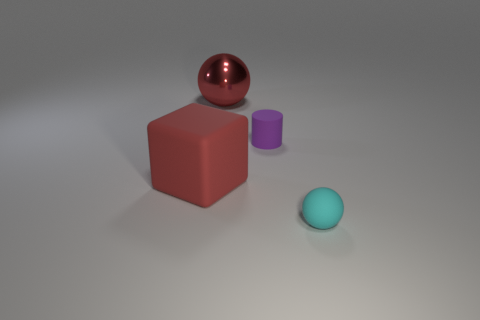}%
  \includegraphics[width=0.49\linewidth]{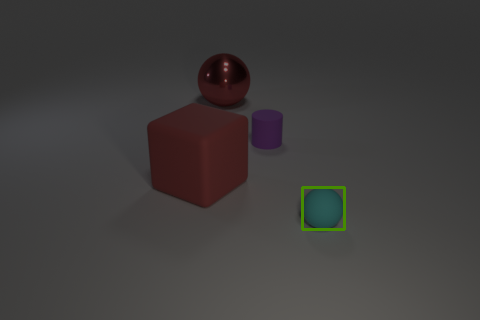}
  \caption{\textit{Look at matte thing that is on the left side of the red object that is behind the second one of the object(s) from right; The first one of the rubber thing(s) from front that are right of it}}

\end{subfigure}&

\begin{subfigure}[t]{0.49\linewidth}
  \centering
  \includegraphics[width=0.49\linewidth]{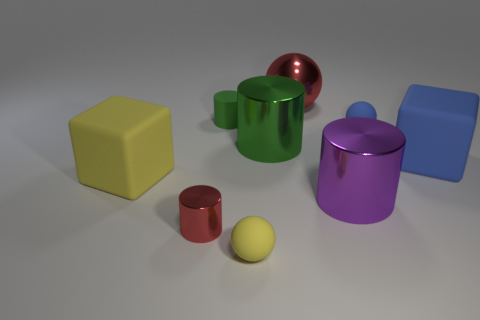}%
  \includegraphics[width=0.49\linewidth]{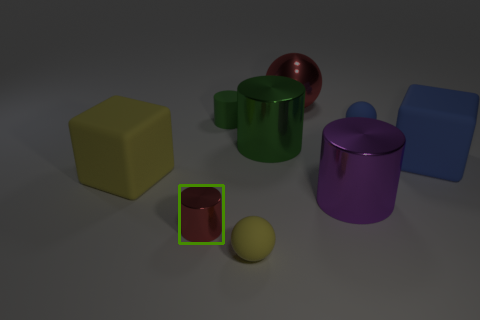}
  \caption{\textit{The objects that are the seventh one of the thing(s) from right that are in front of the nineth one of the thing(s) from front or the second one of the thing(s) from front}}

\end{subfigure}\\

\begin{subfigure}[t]{0.49\linewidth}
  \centering
  \includegraphics[width=0.49\linewidth]{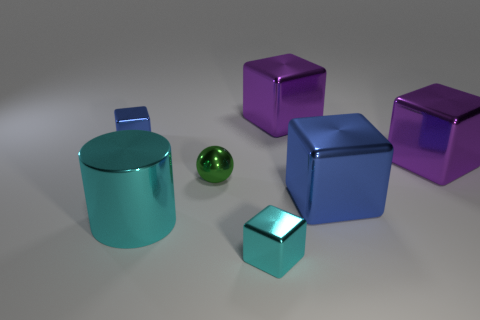}%
  \includegraphics[width=0.49\linewidth]{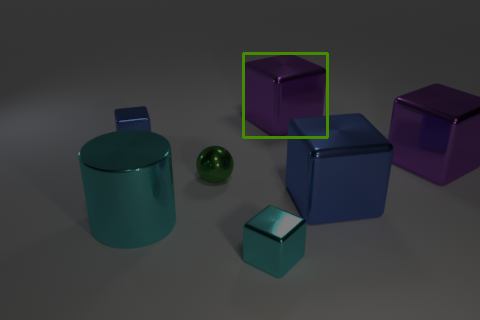}
  \caption{\textit{The big metallic object(s) that are both to the left of the third one of the large thing(s) from left and on the right side of the first one of the object(s) from front}}

\end{subfigure}&

\begin{subfigure}[t]{0.49\linewidth}
  \centering
  \includegraphics[width=0.49\linewidth]{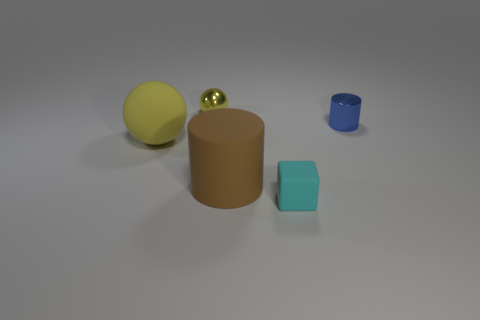}%
  \includegraphics[width=0.49\linewidth]{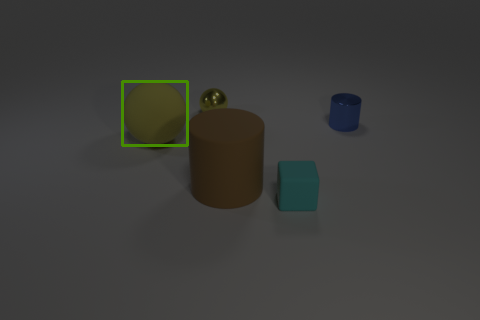}
  \caption{\textit{The fully visible yellow ball(s)}}

\end{subfigure}\\

\begin{subfigure}[t]{0.49\linewidth}
  \centering
  \includegraphics[width=0.49\linewidth]{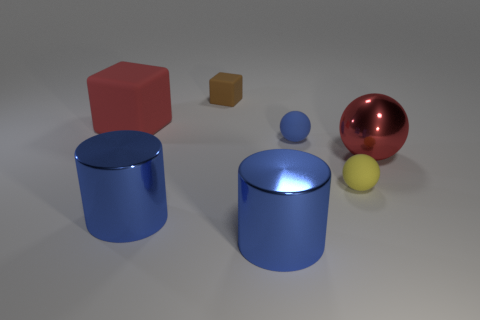}%
  \includegraphics[width=0.49\linewidth]{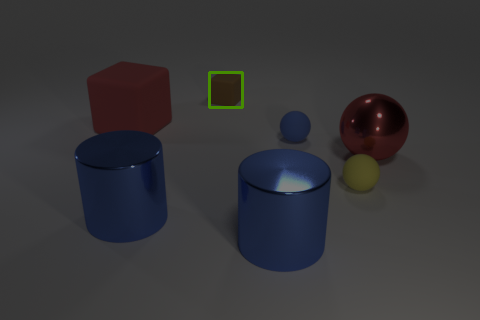}
  \caption{\textit{Any other things that are the same shape as the fourth one of the rubber thing(s) from right}}

\end{subfigure}&

\begin{subfigure}[t]{0.49\linewidth}
  \centering
  \includegraphics[width=0.49\linewidth]{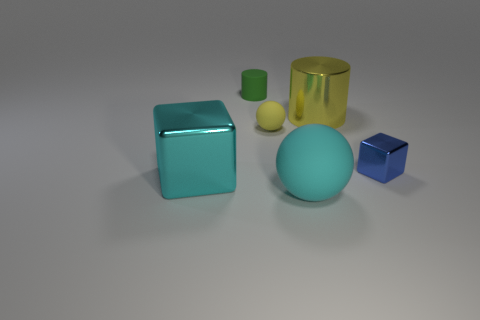}%
  \includegraphics[width=0.49\linewidth]{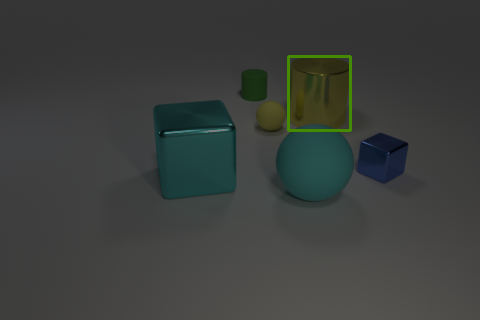}
  \caption{\textit{Find object that is behind the fifth one of the object(s) from left; The cylinder(s) that are to the right of it}}

\end{subfigure}\\

\begin{subfigure}[t]{0.49\linewidth}
  \centering
  \includegraphics[width=0.49\linewidth]{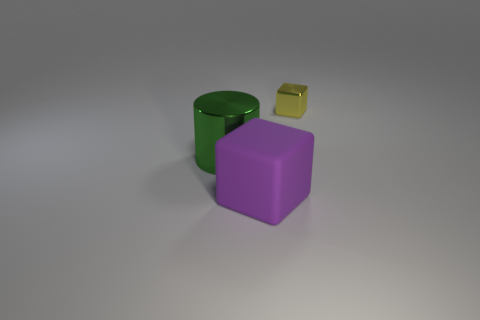}%
  \includegraphics[width=0.49\linewidth]{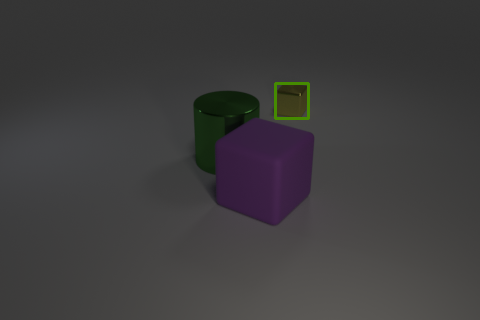}
  \caption{\textit{Look at partially visible object(s); The second one of the thing(s) from left that are on the right side of it}}

\end{subfigure}&

\begin{subfigure}[t]{0.49\linewidth}
  \centering
  \includegraphics[width=0.49\linewidth]{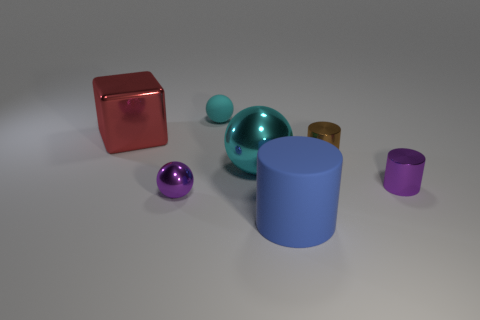}%
  \includegraphics[width=0.49\linewidth]{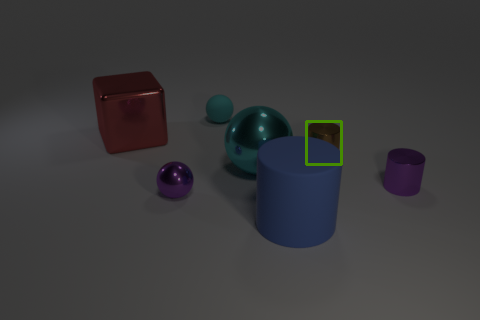}
  \caption{\textit{The second one of the shiny cylinder(s) from right that are to the right of the thing that is behind the thing that is on the left side of the first one of the tiny thing(s) from left}}

\end{subfigure}\\

\begin{subfigure}[t]{0.49\linewidth}
  \centering
  \includegraphics[width=0.49\linewidth]{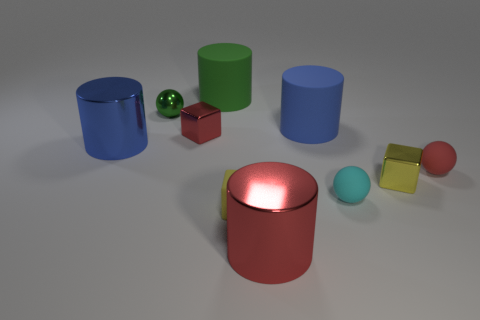}%
  \includegraphics[width=0.49\linewidth]{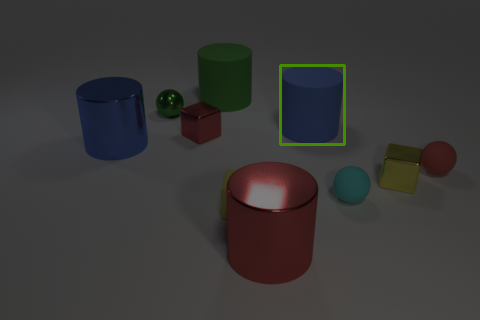}
  \caption{\textit{The blue things that are either the fourth one of the thing(s) from right or the first one of the tiny ball(s) from front}}

\end{subfigure}&

\begin{subfigure}[t]{0.49\linewidth}
  \centering
  \includegraphics[width=0.49\linewidth]{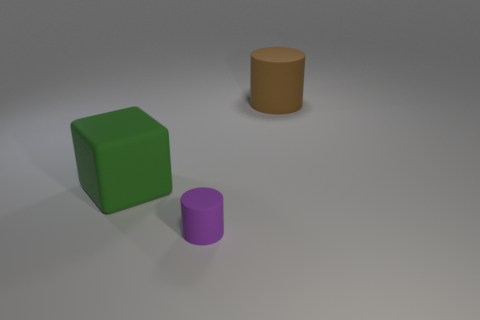}%
  \includegraphics[width=0.49\linewidth]{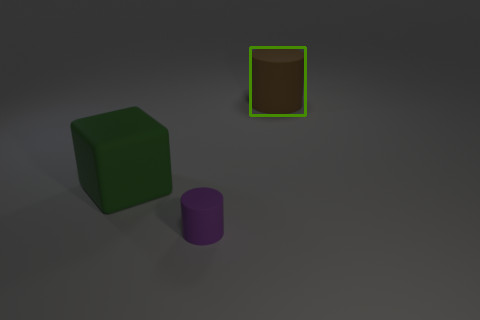}
  \caption{\textit{The matte object(s) that are behind the second one of the cylinder(s) from right and on the right side of the first one of the object(s) from left}}

\end{subfigure}\\

\end{tabular}
\caption{Referring object detection examples from CLEVR-Ref+.}
\label{fig:more-det}
\end{figure*}

\begin{figure*}[hbtp]
\centering
\begin{tabular}[c]{cc}

\begin{subfigure}[t]{0.49\linewidth}
  \centering
  \includegraphics[width=0.49\linewidth]{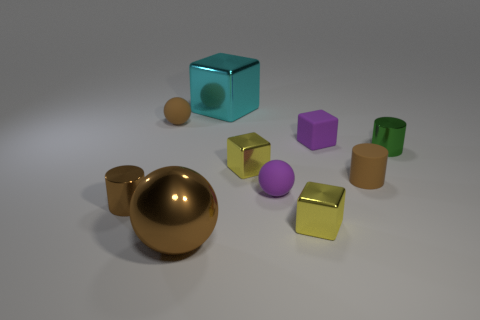}%
  \includegraphics[width=0.49\linewidth]{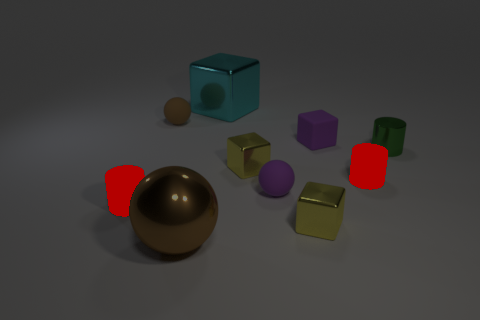}
  \caption{\textit{Any other things that are the same shape as the seventh one of the object(s) from front}}

\end{subfigure}&

\begin{subfigure}[t]{0.49\linewidth}
  \centering
  \includegraphics[width=0.49\linewidth]{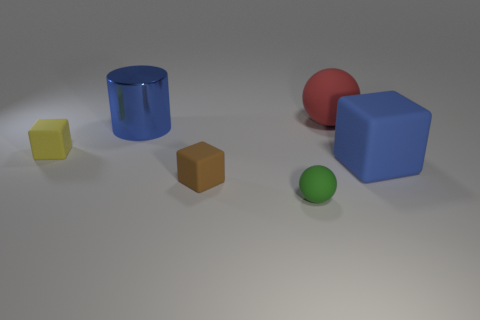}%
  \includegraphics[width=0.49\linewidth]{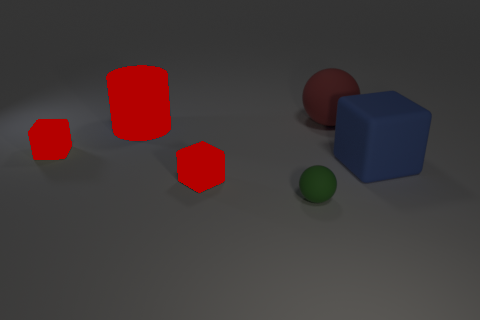}
  \caption{\textit{Look at rubber ball that is to the left of the red ball(s); The thing(s) that are left of it}}

\end{subfigure}\\

\begin{subfigure}[t]{0.49\linewidth}
  \centering
  \includegraphics[width=0.49\linewidth]{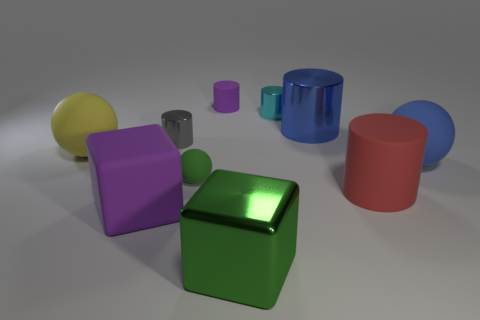}%
  \includegraphics[width=0.49\linewidth]{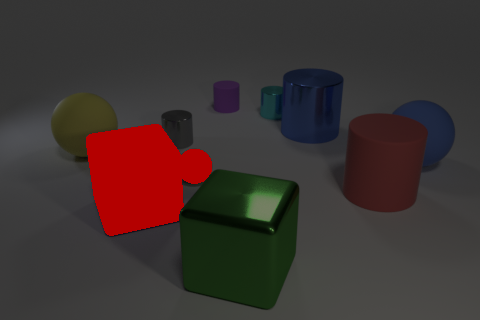}
  \caption{\textit{The rubber object(s) that are to the right of the sixth one of the rubber thing(s) from right and to the left of the fifth one of the object(s) from left}}

\end{subfigure}&

\begin{subfigure}[t]{0.49\linewidth}
  \centering
  \includegraphics[width=0.49\linewidth]{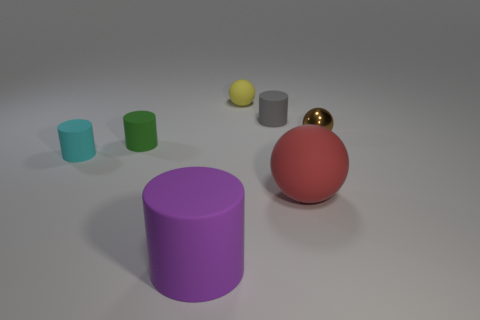}%
  \includegraphics[width=0.49\linewidth]{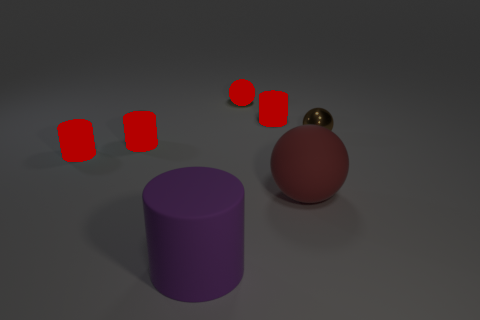}
  \caption{\textit{The fully visible small thing(s)}}

\end{subfigure}\\

\begin{subfigure}[t]{0.49\linewidth}
  \centering
  \includegraphics[width=0.49\linewidth]{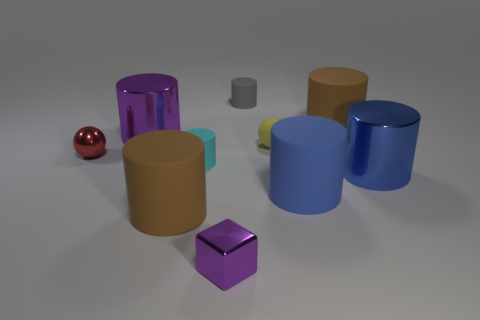}%
  \includegraphics[width=0.49\linewidth]{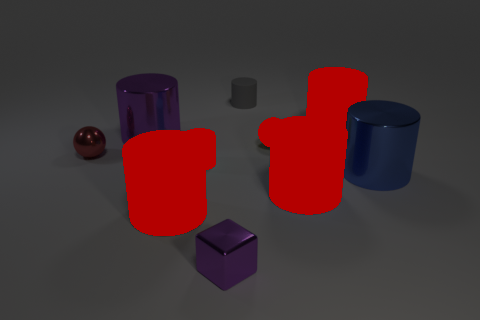}
  \caption{\textit{Look at tiny rubber cylinder that is behind the tiny object that is on the right side of the seventh one of the cylinder(s) from front; The rubber thing(s) that are in front of it}}

\end{subfigure}&

\begin{subfigure}[t]{0.49\linewidth}
  \centering
  \includegraphics[width=0.49\linewidth]{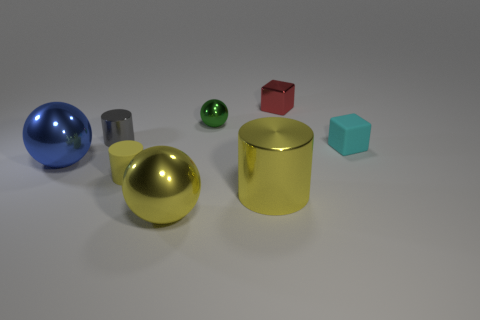}%
  \includegraphics[width=0.49\linewidth]{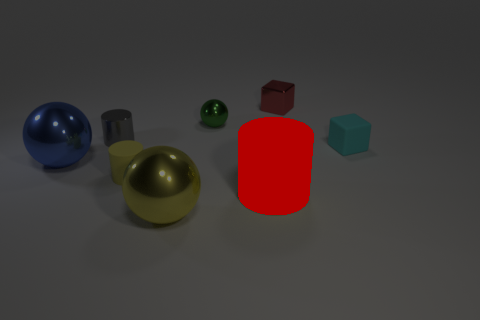}
  \caption{\textit{The big things that are the sixth one of the object(s) from left or the seventh one of the object(s) from right}}

\end{subfigure}\\

\begin{subfigure}[t]{0.49\linewidth}
  \centering
  \includegraphics[width=0.49\linewidth]{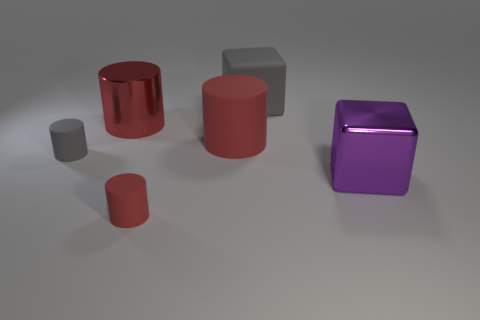}%
  \includegraphics[width=0.49\linewidth]{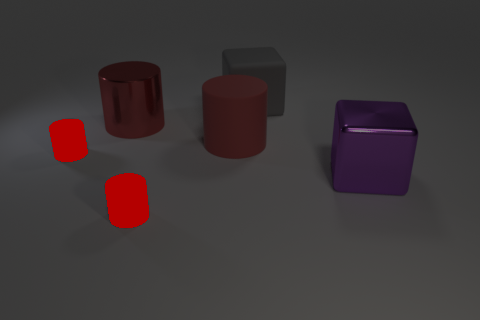}
  \caption{\textit{Find the second one of the red rubber thing(s) from left; The fully visible rubber cylinder(s) that are in front of it}}

\end{subfigure}&

\begin{subfigure}[t]{0.49\linewidth}
  \centering
  \includegraphics[width=0.49\linewidth]{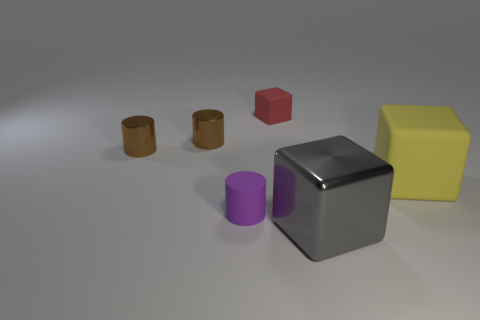}%
  \includegraphics[width=0.49\linewidth]{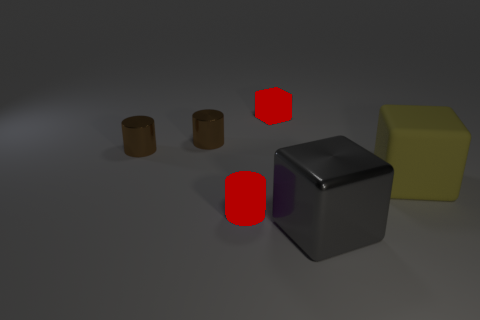}
  \caption{\textit{Any other tiny object(s) made of the same material as the second one of the cube(s) from front}}

\end{subfigure}\\

\begin{subfigure}[t]{0.49\linewidth}
  \centering
  \includegraphics[width=0.49\linewidth]{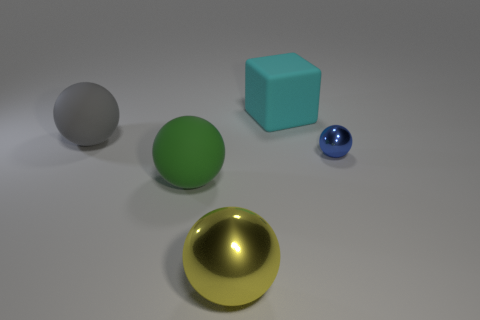}%
  \includegraphics[width=0.49\linewidth]{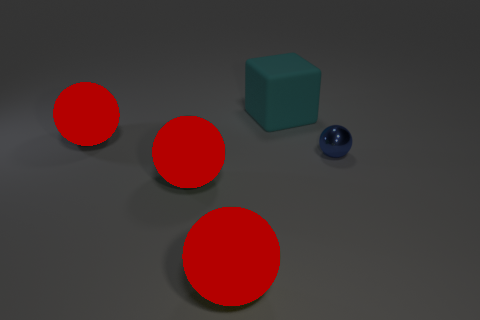}
  \caption{\textit{Look at object that is to the right of the fourth one of the big object(s) from front; The ball(s) that are to the left of it}}

\end{subfigure}&

\begin{subfigure}[t]{0.49\linewidth}
  \centering
  \includegraphics[width=0.49\linewidth]{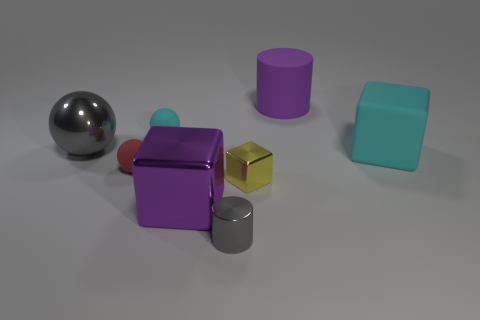}%
  \includegraphics[width=0.49\linewidth]{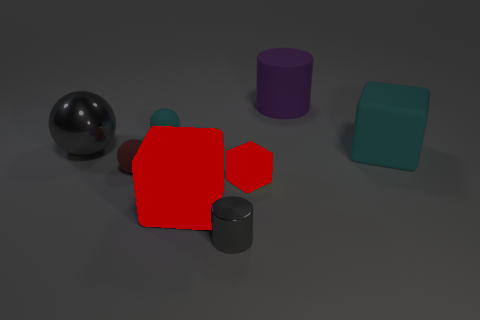}
  \caption{\textit{The metallic object(s) that are behind the fourth one of the object(s) from right and in front of the fourth one of the thing(s) from front}}

\end{subfigure}\\

\end{tabular}
\caption{Referring image segmentation examples from CLEVR-Ref+.}
\label{fig:more-seg}
\end{figure*}

%% file: egpaper_for_review.bbl
\begin{thebibliography}{10}\itemsep=-1pt

\bibitem{DBLP:conf/cvpr/AndreasRDK16}
J.~Andreas, M.~Rohrbach, T.~Darrell, and D.~Klein.
\newblock Neural module networks.
\newblock In {\em {CVPR}}, pages 39--48. {IEEE} Computer Society, 2016.

\bibitem{DBLP:conf/iccv/AntolALMBZP15}
S.~Antol, A.~Agrawal, J.~Lu, M.~Mitchell, D.~Batra, C.~L. Zitnick, and
  D.~Parikh.
\newblock {VQA:} visual question answering.
\newblock In {\em {ICCV}}, pages 2425--2433. {IEEE} Computer Society, 2015.

\bibitem{DBLP:conf/naacl/CirikMB18}
V.~Cirik, L.~Morency, and T.~Berg{-}Kirkpatrick.
\newblock Visual referring expression recognition: What do systems actually
  learn?
\newblock In {\em {NAACL-HLT} {(2)}}, pages 781--787. Association for
  Computational Linguistics, 2018.

\bibitem{DBLP:conf/cvpr/DengDSLL009}
J.~Deng, W.~Dong, R.~Socher, L.~Li, K.~Li, and F.~Li.
\newblock Imagenet: {A} large-scale hierarchical image database.
\newblock In {\em {CVPR}}, pages 248--255. {IEEE} Computer Society, 2009.

\bibitem{DBLP:conf/cvpr/DonahueHGRVDS15}
J.~Donahue, L.~A. Hendricks, S.~Guadarrama, M.~Rohrbach, S.~Venugopalan,
  T.~Darrell, and K.~Saenko.
\newblock Long-term recurrent convolutional networks for visual recognition and
  description.
\newblock In {\em {CVPR}}, pages 2625--2634. {IEEE} Computer Society, 2015.

\bibitem{DBLP:conf/nips/GaoMZHWX15}
H.~Gao, J.~Mao, J.~Zhou, Z.~Huang, L.~Wang, and W.~Xu.
\newblock Are you talking to a machine? dataset and methods for multilingual
  image question.
\newblock In {\em {NIPS}}, pages 2296--2304, 2015.

\bibitem{DBLP:conf/cvpr/GoyalKSBP17}
Y.~Goyal, T.~Khot, D.~Summers{-}Stay, D.~Batra, and D.~Parikh.
\newblock Making the {V} in {VQA} matter: Elevating the role of image
  understanding in visual question answering.
\newblock In {\em {CVPR}}, pages 6325--6334. {IEEE} Computer Society, 2017.

\bibitem{DBLP:conf/cvpr/HeZRS16}
K.~He, X.~Zhang, S.~Ren, and J.~Sun.
\newblock Deep residual learning for image recognition.
\newblock In {\em {CVPR}}, pages 770--778. {IEEE} Computer Society, 2016.

\bibitem{DBLP:conf/eccv/HuADS18}
R.~Hu, J.~Andreas, T.~Darrell, and K.~Saenko.
\newblock Explainable neural computation via stack neural module networks.
\newblock In {\em {ECCV} {(7)}}, volume 11211 of {\em Lecture Notes in Computer
  Science}, pages 55--71. Springer, 2018.

\bibitem{DBLP:conf/iccv/HuARDS17}
R.~Hu, J.~Andreas, M.~Rohrbach, T.~Darrell, and K.~Saenko.
\newblock Learning to reason: End-to-end module networks for visual question
  answering.
\newblock In {\em {ICCV}}, pages 804--813. {IEEE} Computer Society, 2017.

\bibitem{DBLP:conf/cvpr/HuRADS17}
R.~Hu, M.~Rohrbach, J.~Andreas, T.~Darrell, and K.~Saenko.
\newblock Modeling relationships in referential expressions with compositional
  modular networks.
\newblock In {\em {CVPR}}, pages 4418--4427. {IEEE} Computer Society, 2017.

\bibitem{DBLP:conf/eccv/HuRD16}
R.~Hu, M.~Rohrbach, and T.~Darrell.
\newblock Segmentation from natural language expressions.
\newblock In {\em {ECCV} {(1)}}, volume 9905 of {\em Lecture Notes in Computer
  Science}, pages 108--124. Springer, 2016.

\bibitem{DBLP:conf/cvpr/HuXRFSD16}
R.~Hu, H.~Xu, M.~Rohrbach, J.~Feng, K.~Saenko, and T.~Darrell.
\newblock Natural language object retrieval.
\newblock In {\em {CVPR}}, pages 4555--4564. {IEEE} Computer Society, 2016.

\bibitem{hudson2018compositional}
D.~A. Hudson and C.~D. Manning.
\newblock Compositional attention networks for machine reasoning.
\newblock {\em CoRR}, abs/1803.03067, 2018.

\bibitem{DBLP:conf/cvpr/JohnsonHMFZG17}
J.~Johnson, B.~Hariharan, L.~van~der Maaten, L.~Fei{-}Fei, C.~L. Zitnick, and
  R.~B. Girshick.
\newblock {CLEVR:} {A} diagnostic dataset for compositional language and
  elementary visual reasoning.
\newblock In {\em {CVPR}}, pages 1988--1997. {IEEE} Computer Society, 2017.

\bibitem{DBLP:conf/iccv/JohnsonHMHFZG17}
J.~Johnson, B.~Hariharan, L.~van~der Maaten, J.~Hoffman, L.~Fei{-}Fei, C.~L.
  Zitnick, and R.~B. Girshick.
\newblock Inferring and executing programs for visual reasoning.
\newblock In {\em {ICCV}}, pages 3008--3017. {IEEE} Computer Society, 2017.

\bibitem{DBLP:conf/cvpr/KarpathyL15}
A.~Karpathy and F.~Li.
\newblock Deep visual-semantic alignments for generating image descriptions.
\newblock In {\em {CVPR}}, pages 3128--3137. {IEEE} Computer Society, 2015.

\bibitem{DBLP:conf/emnlp/KazemzadehOMB14}
S.~Kazemzadeh, V.~Ordonez, M.~Matten, and T.~L. Berg.
\newblock Referitgame: Referring to objects in photographs of natural scenes.
\newblock In {\em {EMNLP}}, pages 787--798. {ACL}, 2014.

\bibitem{DBLP:journals/corr/KingmaB14}
D.~P. Kingma and J.~Ba.
\newblock Adam: {A} method for stochastic optimization.
\newblock {\em CoRR}, abs/1412.6980, 2014.

\bibitem{li2018referring}
R.~Li, K.~Li, Y.-C. Kuo, M.~Shu, X.~Qi, X.~Shen, and J.~Jia.
\newblock Referring image segmentation via recurrent refinement networks.
\newblock In {\em {CVPR}}, pages 5745--5753. {IEEE} Computer Society, 2018.

\bibitem{DBLP:conf/iccv/LiuLSYLY17}
C.~Liu, Z.~Lin, X.~Shen, J.~Yang, X.~Lu, and A.~L. Yuille.
\newblock Recurrent multimodal interaction for referring image segmentation.
\newblock In {\em {ICCV}}, pages 1280--1289. {IEEE} Computer Society, 2017.

\bibitem{DBLP:conf/aaai/LiuMSY17}
C.~Liu, J.~Mao, F.~Sha, and A.~L. Yuille.
\newblock Attention correctness in neural image captioning.
\newblock In {\em {AAAI}}, pages 4176--4182. {AAAI} Press, 2017.

\bibitem{DBLP:conf/cvpr/LuoS17}
R.~Luo and G.~Shakhnarovich.
\newblock Comprehension-guided referring expressions.
\newblock In {\em {CVPR}}, pages 3125--3134. {IEEE} Computer Society, 2017.

\bibitem{DBLP:conf/cvpr/MaoHTCY016}
J.~Mao, J.~Huang, A.~Toshev, O.~Camburu, A.~L. Yuille, and K.~Murphy.
\newblock Generation and comprehension of unambiguous object descriptions.
\newblock In {\em {CVPR}}, pages 11--20. {IEEE} Computer Society, 2016.

\bibitem{DBLP:journals/corr/MaoXYWY14a}
J.~Mao, W.~Xu, Y.~Yang, J.~Wang, and A.~L. Yuille.
\newblock Deep captioning with multimodal recurrent neural networks (m-rnn).
\newblock {\em CoRR}, abs/1412.6632, 2014.

\bibitem{DBLP:conf/eccv/Margffoy-TuayPB18}
E.~Margffoy{-}Tuay, J.~C. P{\'{e}}rez, E.~Botero, and P.~Arbel{\'{a}}ez.
\newblock Dynamic multimodal instance segmentation guided by natural language
  queries.
\newblock In {\em {ECCV} {(11)}}, volume 11215 of {\em Lecture Notes in
  Computer Science}, pages 656--672. Springer, 2018.

\bibitem{mascharka2018transparency}
D.~Mascharka, P.~Tran, R.~Soklaski, and A.~Majumdar.
\newblock Transparency by design: Closing the gap between performance and
  interpretability in visual reasoning.
\newblock {\em CoRR}, abs/1803.05268, 2018.

\bibitem{DBLP:conf/cvpr/MisraGFHGM18}
I.~Misra, R.~B. Girshick, R.~Fergus, M.~Hebert, A.~Gupta, and L.~van~der
  Maaten.
\newblock Learning by asking questions.
\newblock In {\em {CVPR}}, pages 11--20. {IEEE} Computer Society, 2018.

\bibitem{DBLP:conf/eccv/NagarajaMD16}
V.~K. Nagaraja, V.~I. Morariu, and L.~S. Davis.
\newblock Modeling context between objects for referring expression
  understanding.
\newblock In {\em {ECCV} {(4)}}, volume 9908 of {\em Lecture Notes in Computer
  Science}, pages 792--807. Springer, 2016.

\bibitem{DBLP:conf/aaai/PerezSVDC18}
E.~Perez, F.~Strub, H.~de~Vries, V.~Dumoulin, and A.~C. Courville.
\newblock Film: Visual reasoning with a general conditioning layer.
\newblock In {\em {AAAI}}, pages 3942--3951. {AAAI} Press, 2018.

\bibitem{DBLP:conf/emnlp/RayCBBP16}
A.~Ray, G.~Christie, M.~Bansal, D.~Batra, and D.~Parikh.
\newblock Question relevance in {VQA:} identifying non-visual and false-premise
  questions.
\newblock In {\em {EMNLP}}, pages 919--924. The Association for Computational
  Linguistics, 2016.

\bibitem{DBLP:conf/eccv/RohrbachRHDS16}
A.~Rohrbach, M.~Rohrbach, R.~Hu, T.~Darrell, and B.~Schiele.
\newblock Grounding of textual phrases in images by reconstruction.
\newblock In {\em {ECCV} {(1)}}, volume 9905 of {\em Lecture Notes in Computer
  Science}, pages 817--834. Springer, 2016.

\bibitem{DBLP:conf/nips/SantoroRBMPBL17}
A.~Santoro, D.~Raposo, D.~G.~T. Barrett, M.~Malinowski, R.~Pascanu,
  P.~Battaglia, and T.~Lillicrap.
\newblock A simple neural network module for relational reasoning.
\newblock In {\em {NIPS}}, pages 4974--4983, 2017.

\bibitem{DBLP:conf/cvpr/YuLSYLBB18}
L.~Yu, Z.~Lin, X.~Shen, J.~Yang, X.~Lu, M.~Bansal, and T.~L. Berg.
\newblock Mattnet: Modular attention network for referring expression
  comprehension.
\newblock In {\em {CVPR}}. {IEEE} Computer Society, 2018.

\bibitem{DBLP:conf/eccv/YuPYBB16}
L.~Yu, P.~Poirson, S.~Yang, A.~C. Berg, and T.~L. Berg.
\newblock Modeling context in referring expressions.
\newblock In {\em {ECCV} {(2)}}, volume 9906 of {\em Lecture Notes in Computer
  Science}, pages 69--85. Springer, 2016.

\bibitem{DBLP:conf/cvpr/YuTBB17}
L.~Yu, H.~Tan, M.~Bansal, and T.~L. Berg.
\newblock A joint speaker-listener-reinforcer model for referring expressions.
\newblock In {\em {CVPR}}, pages 3521--3529. {IEEE} Computer Society, 2017.

\bibitem{DBLP:conf/cvpr/ZhangGSBP16}
P.~Zhang, Y.~Goyal, D.~Summers{-}Stay, D.~Batra, and D.~Parikh.
\newblock Yin and yang: Balancing and answering binary visual questions.
\newblock In {\em {CVPR}}, pages 5014--5022. {IEEE} Computer Society, 2016.

\bibitem{DBLP:conf/cvpr/ZhuGBF16}
Y.~Zhu, O.~Groth, M.~S. Bernstein, and L.~Fei{-}Fei.
\newblock Visual7w: Grounded question answering in images.
\newblock In {\em {CVPR}}, pages 4995--5004. {IEEE} Computer Society, 2016.

\end{thebibliography}
